\PassOptionsToPackage{hyphens}{url}
\documentclass{article}
\usepackage{iclr2026_conference,times}

\usepackage{amsmath,amsfonts,bm}

\def\eqref#1{equation~\ref{#1}}

\def\1{\bm{1}}

\DeclareMathAlphabet{\mathsfit}{\encodingdefault}{\sfdefault}{m}{sl}
\SetMathAlphabet{\mathsfit}{bold}{\encodingdefault}{\sfdefault}{bx}{n}

\usepackage{hyperref}
\usepackage{url}
\usepackage{graphicx}
\usepackage{amsmath}
\usepackage{amssymb}
\usepackage{booktabs}
\usepackage{multirow}
\usepackage{microtype}
\usepackage{dblfloatfix}
\usepackage{capt-of}

\newcommand{\FSD}{\mathrm{FSD}}
\newcommand{\MDL}{\mathrm{MDL}}
\newcommand{\JS}{\mathrm{JS}}
\newcommand{\SG}{\mathrm{SG}}
\newcommand{\suppfig}[1]{Supplementary Fig.~S#1}
\newcommand{\supptab}[1]{Supplementary Table~S#1}
\newcommand{\cmark}{\checkmark}

\title{Hidden APIs in Language Models: Discovering Reusable Causal Interfaces from Forked Futures}

\author{\parbox{0.96\textwidth}{\centering
SiYuan Ma$^{1}$,
Yiqin Luo$^{2}$,
Zhangji$^{3}$,
Canran Xiao$^{4}$,
Albert Gao$^{5}$\\
Wei Wang$^{6}$,
Qiwei Wu$^{7}$,
Xinran Li$^{8}$\\
Jinfeng Wei$^{9}$,
Qixin Zhang$^{1}$\\[0.7em]
{\small
$^{1}$Nanyang Technological University
\quad
$^{2}$Southern University of Science and Technology\\
$^{3}$Tianjin University
\quad
$^{4}$Sun Yat-sen University\\
$^{5}$Carnegie Mellon University
\quad
$^{6}$The Hong Kong Polytechnic University\\
$^{7}$Shenzhen University
\quad
$^{8}$Tsinghua University\\
$^{9}$The Hong Kong University of Science and Technology}\\[0.5em]
\texttt{masi0004@e.ntu.edu.sg}}}

\iclrfinalcopy

\begin{document}
\maketitle

\begin{abstract}
Identical language-model answers can arise from hidden states that support different future computations, so current-answer probes do not establish a reusable internal interface. We introduce \emph{forked futures}: future operations are sampled only after a prefix state has formed, and states are compared through the response distributions induced by those operations. This yields an empirical causal quotient over hidden states without requiring researcher-specified latent labels. Shared, Local, Mixture, and Distributed interfaces then compete under prequential causal description length subject to future-signature fidelity and matched capacity constraints. In the two detailed model evaluations, Shared has the lowest held-out description length, with gains of 0.216 nats on Qwen2.5-1.5B and 0.294 nats on Llama-3-8B, while maintaining tightly clustered mean future-signature distortion; a five-backbone sweep preserves the positive direction of Sharedness Gain. The figure-aligned transplantation analysis gives Shared the strongest joint target-correctness, locality, copy-preservation, and composite profile, and API-aligned paths mediate 0.749 of the target effect versus 0.150 for matched null paths. In the blind four-class model-organism test, 14/16 architectures are recovered, with one observed non-Shared $\rightarrow$ Shared error among 12 non-Shared organisms. These results support an economical reusable causal interface within the tested operation banks, while keeping the claim explicitly conditional on the candidate architectures, interventions, and held-out futures.
\end{abstract}

\section{Introduction}

A hidden state can answer the current question correctly for incompatible reasons. Two prefixes may emit the same token yet respond differently when later asked to compare, verify, invert, execute, or compose the underlying state. Decodability from a single answer therefore does not establish a shared computational interface \cite{16,17}. The sharper question is counterfactual: \emph{which distinctions between hidden states remain necessary across possible future computations?}

Global-workspace theories suggest that specialized processes communicate through a compact shared state \cite{2,3,4,5,6,7}, and recent work reports globally available verbalizable representations in language models \cite{1}. These observations motivate a mechanistic hypothesis but do not distinguish a shared interface from operation-specific or distributed alternatives. We address this gap with forked futures (conceptual overview in \suppfig{6}). For a prefix state $h$, an operation bank $\mathcal K$ generates valid continuations, and the resulting response distributions define a future causal signature. Compact encoders of this signature are compared as Shared, Local, Mixture, or Distributed using prequential causal MDL \cite{14,15}. The selected interface must then pass held-out sufficiency, transplantation, mediation, matched-baseline, and ground-truth falsification tests. Full data-construction controls and evaluation criteria are reported in \suppfig{1}, \supptab{1}, and \supptab{11}.

The contribution is a single evidence chain: future-defined causal equivalence prevents current-answer shortcuts; explicit architecture competition controls complexity; role-aligned interventions test reuse and causal consumption; and four model-organism classes measure false Shared discoveries. Each component addresses a distinct failure mode. A decodable direction may be epiphenomenal, a low-rank code may win only because it is cheaper, a successful transplant may copy donor content, and a sparse path map may reflect patchability rather than mediation. We therefore treat a Shared interface as supported only when these alternative explanations are jointly constrained.

The resulting claim is deliberately narrower than identifying a unique global workspace for all language-model computation. We ask whether, for a preregistered family of future operations, the model exposes a compact state that multiple downstream computations can reuse. The claim is comparative: Shared is accepted only relative to explicit Local, Mixture, and Distributed alternatives under matched capacity and held-out fidelity. Additional sensitivity analyses, statistical tests, and reproducibility details are provided in the supplementary material.

\section{Related Work}

Global workspace theories model capacity-limited broadcast among specialized modules \cite{2,3,4,5}, and machine-learning variants coordinate modular networks through a common workspace \cite{6,7}. These accounts motivate the possibility of a reusable internal state, but global decodability or verbal report alone does not determine whether several computations actually route through the same interface. Our protocol therefore treats Shared as one architecture class in a falsifiable competition rather than as a default interpretation.

Causal abstraction formalizes when a low-level system implements a higher-level causal model \cite{8,9,10,11}. Interchange interventions, path patching, Patchscopes, and automated circuit discovery test whether internal components causally support behavior \cite{10,18,19,20}. We extend this logic from one output or one circuit to a bank of future operations: the abstraction must preserve distributions over unseen futures, and the same code must be selectively consumed by multiple downstream roles.

Representation methods including linear probes, representation engineering, activation addition, task vectors, sparse autoencoders, transcoders, PCA, and ICA show that low-dimensional activation structure can encode or steer behavior \cite{21,22,23,24,25,26,27,28,29,30,31}. These methods form strong matched baselines, but encoding or steering does not by itself imply a reusable interface. Our criterion jointly requires future sufficiency, short description length, locality-preserving transplantation, selective necessity, and causal mediation. Finally, distribution-shift benchmarks \cite{34,35,36} show that interpolation can overstate systematic generalization, while model-organism studies \cite{37,38,39,40} show that mechanistic detectability depends on training construction. These observations motivate held-out compositions, generator and natural-family shifts, full class confusion, and explicit false-Shared rates.

\section{Future-Defined Causal Interfaces}

Let $x_{1:t}$ be a prefix and $h=h_{\ell,t}(x)$ its hidden state before a future operation is revealed. Operation $k\in\mathcal K$ induces
\begin{equation}
    \sigma_k(h)=p_\theta(y\mid x_{1:t},k),
\end{equation}
and the future causal signature is
\begin{equation}
    \Sigma_{\mathcal K}(h)=\{\sigma_k(h):k\in\mathcal K\}.
    \label{eq:signature}
\end{equation}
We use full-vocabulary logits by default and test top-$100$, answer-token, and sequence log-likelihood variants (\suppfig{2}; \supptab{3}). Two states are empirically future-equivalent when their signatures are close under
\begin{equation}
 \FSD(h,\hat h)=
 \mathbb E_{k\sim\pi_{\mathcal K}}
 \left[\JS\!\left(\sigma_k(h)\,\|\,\sigma_k(\hat h)\right)\right].
 \label{eq:fsd}
\end{equation}
This separates prefixes that share a present answer but diverge after a later operation; \suppfig{1} gives semantic-collision and lexical anti-collision controls. In the semantic-collision condition, two states are matched on the current answer but constructed to differ in their valid future transitions. In the lexical anti-collision condition, surface forms are varied while the latent state and future signature are held fixed. Together, these controls make it difficult for an interface to succeed by memorizing the current answer, prompt template, entity names, or operation labels (\suppfig{1}; \supptab{13}).

Each candidate contains an encoder and reconstructor,
\begin{equation}
    c=E_\phi(h,k),\qquad \hat h=R_\psi(c,k).
\end{equation}
Shared uses one operation-agnostic code $E_\phi(h)$ and allows operation identity to enter only at reconstruction or consumption. Local uses independent $(E_k,R_k)$ pairs and therefore permits every future operation to maintain its own private state. Mixture interpolates between shared and local components through an operation-conditioned gate. Distributed retains the signature across several components without assigning a privileged bottleneck. These alternatives cover the principal ways a compact code can appear shared without actually being reused. Rank, sparsity, total trainable parameters, optimization budget, and the number of prequential updates are matched wherever the architecture permits; unavoidable excess capacity in Distributed is charged by description length rather than ignored (\supptab{2}; \supptab{14}).

\begin{figure*}[!t]
    \centering
    \includegraphics[width=0.90\textwidth]{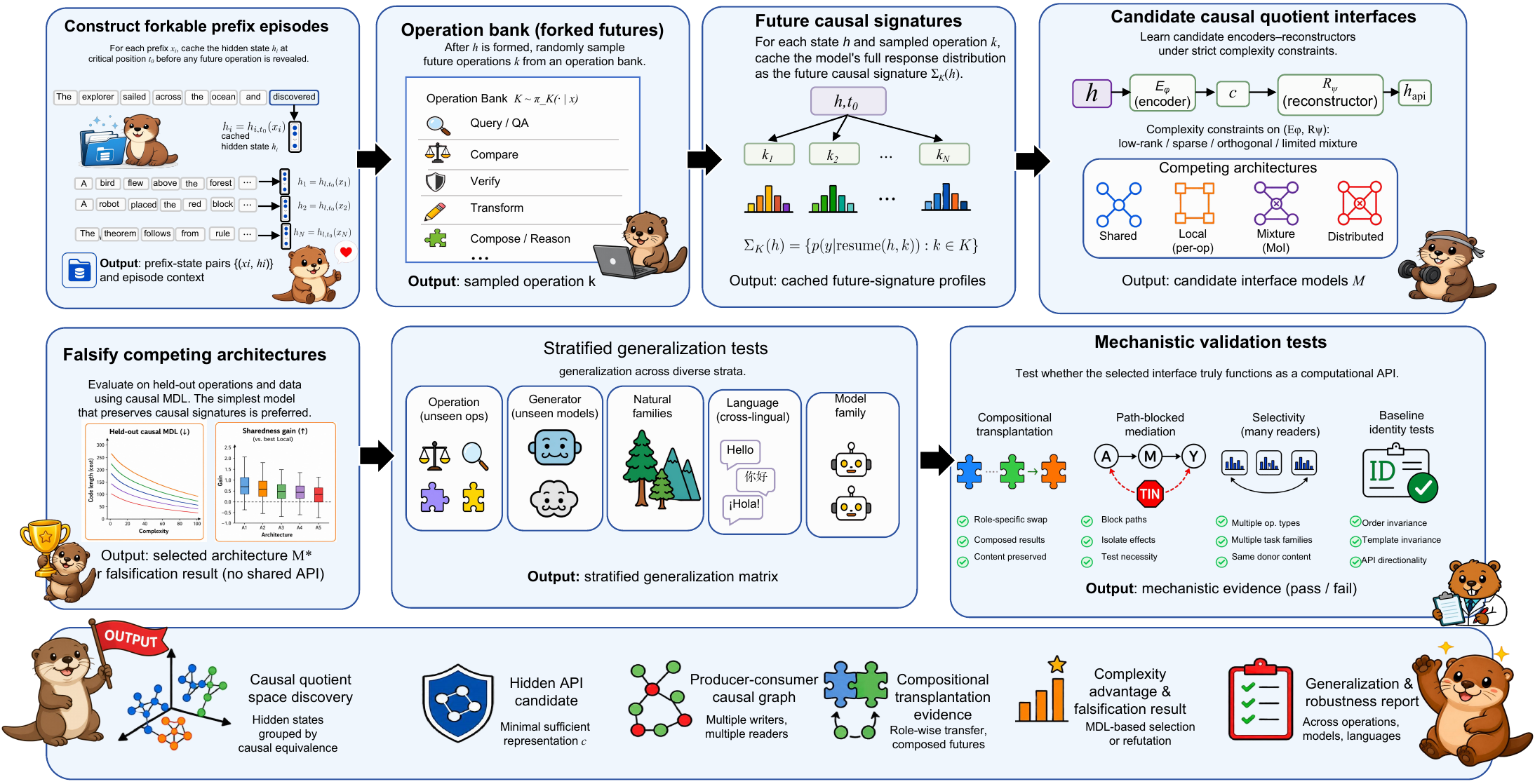}
    \caption{Discovery and falsification pipeline. (A--C) Prefix states are cached before future operations and converted into distribution-valued signatures. (D) Shared, Local, Mixture, and Distributed compete under causal MDL and worst-family FSD. (E) The selected interface is tested by OOD, transplantation, mediation, and identity controls.}
    \label{fig:pipeline}
\end{figure*}

Figure~\ref{fig:pipeline} separates discovery from validation. Panels A--B enforce the central temporal control: the prefix state is cached before the future operation is sampled, preventing operation-name leakage. Panel C converts the resulting continuations into distribution-valued future signatures. Panel D then compares Shared, Local, Mixture, and Distributed under the same fidelity and complexity contract, rather than evaluating Shared in isolation. Panel E is not used to fit the interface; it tests unseen operations, generators, families, interventions, and representation controls. Thus the pipeline first identifies the shortest future-sufficient code and only afterward asks whether the original model reuses it causally (full configurations in \supptab{1}, \supptab{13}, and \supptab{14}).
For ordered blocks $\mathcal D_1,\ldots,\mathcal D_B$,
the prequential causal MDL is
\begin{equation}
\begin{split}
\MDL_{\mathrm{causal}}(A)
={}&L_0(\mathcal D_1) \\
&+\sum_{b=2}^{B}
-\log p_{\hat{\eta}_{A,<b}}
\bigl(\Sigma_b \mid h_b,k_b\bigr).
\end{split}
\label{eq:mdl}
\end{equation}
Here, $\hat{\eta}_{A,<b}$ is trained only on the preceding
blocks. We select
\begin{equation}
A^\star
=
\underset{
\substack{
A:\\
\max_f \FSD_f(A)\leq\epsilon_f
}
}{\arg\min}
\MDL_{\mathrm{causal}}(A).
\label{eq:architecture_selection}
\end{equation}

Let
$\mathcal A_{\mathrm{alt}}
=\{\mathrm{Local},\mathrm{Mixture},\mathrm{Distributed}\}$.
The Sharedness Gain is
\begin{equation}
\begin{split}
\SG
={}&
\min_{A\in\mathcal A_{\mathrm{alt}}}
\MDL_{\mathrm{causal}}(A) \\
&-
\MDL_{\mathrm{causal}}(\mathrm{Shared}).
\end{split}
\label{eq:sharedness_gain}
\end{equation}
Positive $\SG$ favors Shared only when the fidelity constraint also holds. We report mean, worst-family, and 95th-percentile FSD so that a short code cannot win by discarding one difficult family. Layer and rank are selected on validation data and frozen before testing, while uncertainty is computed with family-level paired bootstrap units and matched-label permutation tests. The complete layer, tail, and statistical diagnostics are reported in \suppfig{3} and \supptab{7}--\supptab{8}.

\section{Mechanistic Validation}

For each transplant, donor and base episodes are aligned at the same producer role. The same-family condition also matches task family, difficulty bin, output length, insertion layer and position, and code norm; every control reuses the same base episodes. Cross-family donors alter family-level semantics, matched-random donors preserve intervention load without causal identity, and Local-API donors insert the operation-specific code selected by the Local architecture. This design makes the donor condition, rather than example difficulty or perturbation scale, the primary difference between comparisons (complete matrices in \suppfig{4} and \supptab{4}).

Target Correctness (TC) is the fraction of interventions that realize the task-defined donor-side relation or state update. Locality (LOC) is the fraction of pre-specified non-target slots whose base state is preserved. The third bar in Fig.~\ref{fig:transplant}C is a higher-is-better copy-preservation score,
\begin{equation}
 \mathrm{CopyPres}=1-\mathrm{DonorCopyRate},
\end{equation}
where DonorCopyRate counts unnecessary donor-specific entities, register values, premise tokens, or other surface content. To use the same orientation for all displayed metrics, the figure reports
\begin{equation}
 \mathrm{CompScore}=\frac{\mathrm{TC}+\mathrm{LOC}+\mathrm{CopyPres}}{3}.
 \label{eq:compscore}
\end{equation}
The three components are also reported separately so that the composite cannot hide broad overwrite or donor-content leakage. Operation consistency further evaluates whether one donor code can be consumed by several unseen operations rather than only the operation used to form the pair (\supptab{4}).

For a target effect $\Delta_{\mathrm{total}}$, blocking an API-aligned path gives
\begin{equation}
 \mathrm{MF}=
 \frac{\Delta_{\mathrm{total}}-\Delta_{\mathrm{blocked}}}
      {\Delta_{\mathrm{total}}}.
\end{equation}
Mediated fraction, necessity--restoration, and multi-consumer blocking are compared with null paths matched by producer--consumer distance, rank, activation norm, intervention load, and baseline patchability. We report both thresholded graph statistics and threshold-free precision--recall summaries, so the mechanism cannot be attributed to one favorable cutoff (full path and threshold analyses in \suppfig{4}--\suppfig{5} and \supptab{10}).

The ground-truth organisms are six-layer transformers with hidden width 256 and vocabulary size 512, trained for 3,000 steps with five seeds per architecture. Shared routes all roles through one bottleneck; Local assigns independent workspaces; Mixture gates shared and local routes; and Distributed spreads the relevant variables over components without a privileged bottleneck. The pipeline observes only activations and future signatures, not the class label, making non-Shared $\rightarrow$ Shared errors the primary falsification endpoint (seed-level records in \suppfig{2} and \supptab{5}).

\begin{figure*}[!t]
    \centering
    \includegraphics[width=0.88\textwidth]{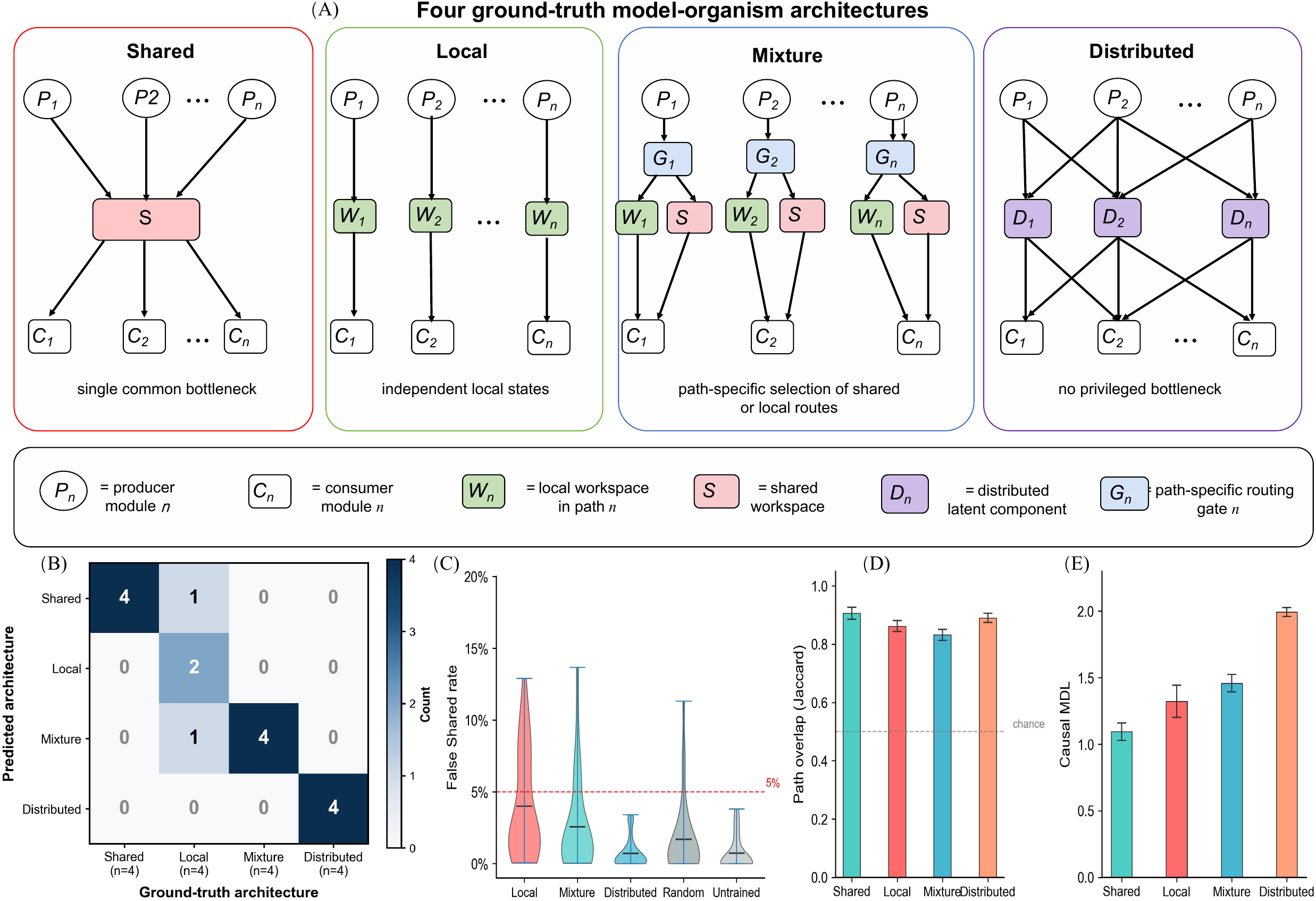}
    \caption{Ground-truth falsification. (A) Four known routing architectures. (B) Blind recovery confusion. (C) False-Shared rates against the 5\% tolerance. (D) Path overlap with the known graph. (E) Class-wise causal MDL.}
    \label{fig:organisms}
\end{figure*}

Figure~\ref{fig:organisms} evaluates whether the discovery rule prefers Shared by construction. Panel A specifies the four routing mechanisms. Panel B contains four held-out organisms per class and recovers 14/16 labels. Shared, Mixture, and Distributed are each recovered in all four cases; among the four Local organisms, two are recovered as Local, one is labeled Shared, and one is labeled Mixture. Panel C therefore records one observed false-Shared error among 12 non-Shared organisms (8.3\%), rather than a zero false-positive result. Panel D shows high overlap with the known communication graphs, and panel E shows clear class-wise MDL ordering. The experiment still demonstrates nontrivial four-class discrimination, but the visible Local $\rightarrow$ Shared error prevents claiming that the falsification tolerance is fully satisfied (full confusion and interpretation in \supptab{5}).

\section{Experiments}

\begin{figure*}[!t]
    \centering
    \includegraphics[width=0.90\textwidth]{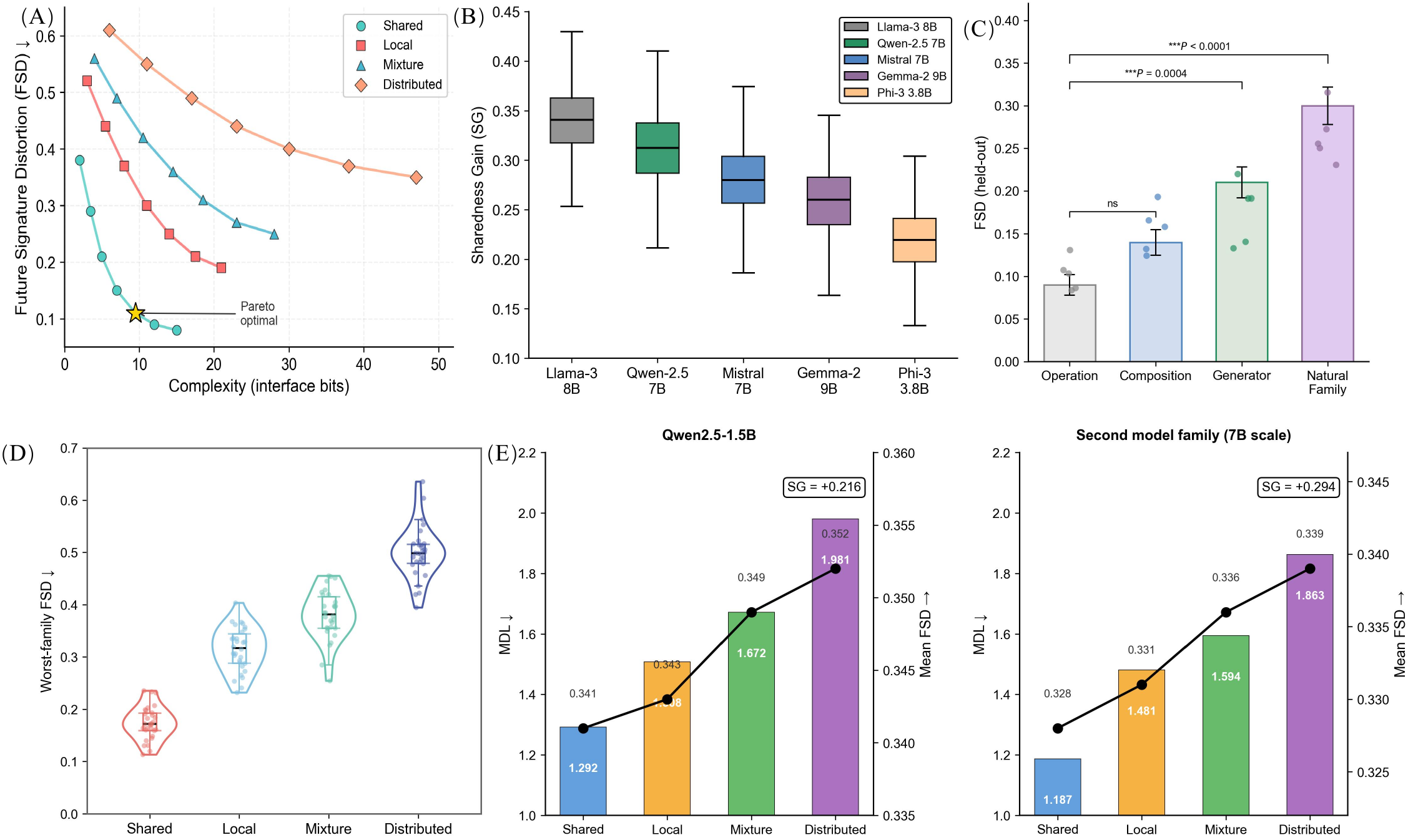}
    \caption{Architecture competition. (A) Complexity--FSD frontiers.
    (B) Sharedness Gain across models. (C) OOD degradation.
    (D) Worst-family FSD. (E) Joint MDL/FSD comparisons on Qwen and
    Llama.}
    \label{fig:competition}
\end{figure*}

We conduct the full architecture, family-stratified, and intervention analyses on Qwen2.5-1.5B and Llama-3-8B \cite{42}. Fig.~\ref{fig:competition}B additionally reports a directional Sharedness-Gain robustness sweep on Qwen2.5-7B, Mistral-7B, Gemma-2-9B, and Phi-3-3.8B. These four additional backbones are used only to test whether the sign of the architecture preference transfers beyond the two fully evaluated models; all family-level, transplantation, mediation, and representation-identity conclusions are based on Qwen2.5-1.5B and Llama-3-8B.

The detailed evaluation uses three structured families. The relation-world split contains 4,200/900/1,200 train/validation/test prefix episodes; program state contains 3,800/820/1,080; and proof/discourse state contains 2,600/560/740, for 10,600 training, 2,280 validation, and 3,020 test episodes in total. All future branches generated from one latent prefix remain in the same split. Relation operations include querying, comparing, inverting, and composing graph relations; program operations include executing, verifying, copying, and updating registers; proof/discourse operations include entailment queries, verification, alternate derivations, and chained composition.

Each family supplies 16 interface-training operations and four unseen operations, together with unseen two- and three-step compositions. Six template families are crossed with four entity/relation pools, and two complete template families are reserved for generator OOD evaluation. Five interface seeds are crossed with five prompt-generation seeds per model. Layers, ranks, thresholds, and fidelity margins are selected on validation data and then frozen. Natural-family evaluation trains on two families and evaluates the third without refitting (\supptab{1}; \supptab{13}).

\begin{table*}[!t]
\centering
\small
\setlength{\tabcolsep}{4pt}
\renewcommand{\arraystretch}{0.96}
\begin{tabular}{p{0.16\textwidth}p{0.79\textwidth}}
\toprule
\textbf{Dimension} & \textbf{Configuration} \\
\midrule

Detailed models
& Qwen2.5-1.5B ($d=1536$, 28 layers);
Llama-3-8B ($d=4096$, 32 layers). \\

Additional SG sweep
& Qwen2.5-7B, Mistral-7B, Gemma-2-9B, and Phi-3-3.8B;
used only for directional backbone robustness. \\

Data
& Relation: 4,200/900/1,200;
Program: 3,800/820/1,080;
Proof/discourse: 2,600/560/740
train/validation/test episodes. \\

Operations
& 16 training and 4 held-out operations per family;
query, compare, invert, compose, execute, verify, update,
entail, and chain; unseen 2--3 step compositions. \\

OOD protocol
& Six template families $\times$ four entity/relation pools;
two template families held out;
operation, composition, generator, and family shifts. \\

Evaluation
& Five interface $\times$ five prompt seeds;
paired family-level bootstrap, permutation, mixed-effects,
TOST, non-inferiority, and BH-FDR. \\

\bottomrule
\end{tabular}
\caption{Experimental data and protocol. The additional four backbones
are used only for the Sharedness-Gain sweep; the complete intervention
and family-stratified analyses use Qwen2.5-1.5B and Llama-3-8B.
Detailed generator, optimization, and compute settings are reported in
\supptab{13}--\supptab{14}.}
\label{tab:scope_main}
\end{table*}

The causal code uses 20 sequential prequential folds. For every later fold, the data term is the cross-entropy code of held-out future-signature distributions in nats. Model complexity is transmitted using a Gaussian-prior two-part parameter code together with the explicit penalty $\lambda_r r+\lambda_g g$, where $r$ is active rank, $g$ is the number of gates, and $(\lambda_r,\lambda_g)=(0.02,0.05)$. Active consumer codes use rank 64, while parameter-matched variants adjust their internal widths. Shared, Local, Mixture, and Distributed instantiate 1, 6, 4, and 16 interface components, respectively. Within each seed, all candidates reuse the same fold ordering, held-out signatures, optimizer schedule, stopping rule, and validation-selected configuration (\supptab{14}).

Table~\ref{tab:main_architecture} gives the primary architecture comparison. On Qwen, Shared reduces MDL from 1.508 for the nearest alternative, Local, to 1.292, yielding $\SG=0.216$ with 95\% CI $[0.161,0.278]$. On Llama, the corresponding reduction is 1.481 to 1.187, with $\SG=0.294$ and CI $[0.208,0.391]$. Mean FSD changes by at most 0.011 within either model, and Shared is also slightly better on worst-family and tail FSD. The result is therefore a lower coding cost at matched predictive adequacy, rather than a large reconstruction-accuracy gap.

\begin{table*}[!t]
\centering
\small
\setlength{\tabcolsep}{3.5pt}
\renewcommand{\arraystretch}{0.94}
\begin{tabular*}{\textwidth}{
@{\extracolsep{\fill}}
llccccc
@{}
}
\toprule
\textbf{Model}
& \textbf{Architecture}
& \textbf{MDL}
& \textbf{Mean FSD}
& \shortstack{\textbf{Worst}\\\textbf{FSD}}
& \shortstack{\textbf{Tail$_{95}$}\\\textbf{FSD}}
& \shortstack{\textbf{SG}\\\textbf{[95\% CI]}} \\
\midrule

\multirow{4}{*}{Qwen2.5-1.5B}
& \textbf{Shared}
& \textbf{1.292}
& \textbf{0.341}
& \textbf{0.358}
& \textbf{0.372}
& \multirow{4}{*}{$0.216\ [0.161,0.278]$} \\

& Local
& 1.508
& 0.343
& 0.361
& 0.379
& \\

& Mixture
& 1.672
& 0.349
& 0.367
& 0.385
& \\

& Distributed
& 1.981
& 0.352
& 0.371
& 0.392
& \\

\midrule

\multirow{4}{*}{Llama-3-8B}
& \textbf{Shared}
& \textbf{1.187}
& \textbf{0.328}
& \textbf{0.344}
& \textbf{0.361}
& \multirow{4}{*}{$0.294\ [0.208,0.391]$} \\

& Local
& 1.481
& 0.331
& 0.349
& 0.368
& \\

& Mixture
& 1.594
& 0.336
& 0.358
& 0.374
& \\

& Distributed
& 1.863
& 0.339
& 0.362
& 0.381
& \\

\bottomrule
\end{tabular*}
\caption{Held-out architecture competition. Lower values are better for
MDL and FSD. SG is the MDL advantage of Shared over the best
non-Shared alternative. Mean, worst-family, and tail distortion are
reported separately so that a short description cannot win by
discarding one difficult family.}
\label{tab:main_architecture}
\end{table*}

Figure~\ref{fig:competition} shows where the MDL advantage arises and where it weakens. Panel A places Shared on the low-complexity portion of the complexity--FSD frontier rather than at a high-distortion corner. Panel B shows positive Sharedness Gain across five backbones, while the complete architecture and uncertainty analyses are reserved for the two detailed models. Panel C orders the distribution shifts: held-out operations are mildest, followed by unseen compositions and generators, while natural-family transfer is hardest. Panel D reports family-wise worst-case distributions and preserves the Shared, Local, Mixture, and Distributed ordering. Panel E reproduces the exact MDL and mean-FSD values in Table~\ref{tab:main_architecture}: the architectures are tightly clustered in measured fidelity but widely separated in description length. Signature, temperature, top-$k$, rank, and layer-position diagnostics are reported in \suppfig{2}--\suppfig{3} and \supptab{3}, \supptab{6}--\supptab{8}.

\begin{figure*}[!t]
    \centering
    \includegraphics[width=0.90\textwidth]{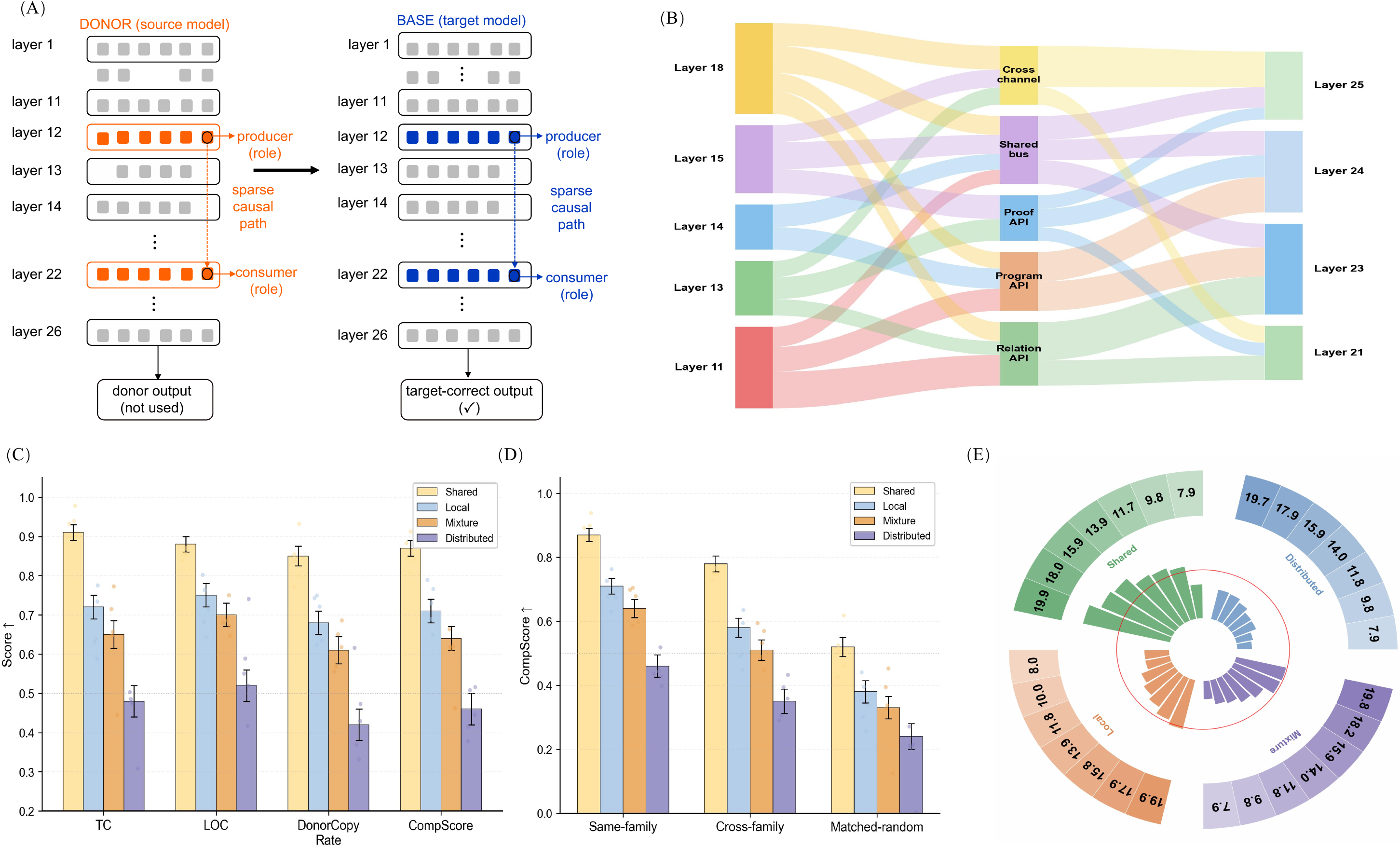}
    \caption{Transplantation and mediation. (A) Role-aligned code
    insertion. (B) Recovered producer--consumer paths. (C) Target
    correctness, locality, copy preservation, and CompScore.
    (D) Donor controls. (E) Sparse path-level causal effects.}
    \label{fig:transplant}
\end{figure*}

To make the quantities in Fig.~\ref{fig:transplant}C directly comparable, donor-copy rate is converted to the higher-is-better quantity
$\mathrm{CopyPres}=1-\mathrm{DonorCopy}$. The figure-level composite is
$\mathrm{CompScore}=(\mathrm{TC}+\mathrm{LOC}+\mathrm{CopyPres})/3$.
The displayed bars are seed-averaged summaries rounded to two decimal places; statistical comparisons use the corresponding unrounded constituent measurements.

\begin{center}
\begin{minipage}{0.98\columnwidth}
\centering
\captionof{table}{Transplantation, mediation, and selectivity summary.
Main-figure values are rounded to two decimals. Complete intervention,
operation-level, and matched-null analyses are reported in
\supptab{4}, \supptab{9}, and \suppfig{4}.}
\label{tab:mechanistic_summary}

\vspace{-1mm}
\scriptsize
\setlength{\tabcolsep}{1.5pt}
\renewcommand{\arraystretch}{0.91}
\begin{tabular}{
@{}
p{0.22\columnwidth}
p{0.72\columnwidth}
@{}
}
\toprule
\textbf{Evidence}
&
\textbf{Figure-aligned result / comparison}
\\
\midrule

Panel-C metrics
&
Shared/Local/Mixture/Distributed:
TC $0.91/0.72/0.65/0.48$;
LOC $0.88/0.75/0.69/0.52$;
CopyPres $0.85/0.66/0.61/0.42$;
CompScore $0.87/0.70/0.64/0.46$.
\\

Donor controls
&
CompScore for same-family
$0.88/0.71/0.64/0.47$;
cross-family
$0.78/0.58/0.51/0.35$;
matched-random
$0.53/0.39/0.34/0.24$
(Shared/Local/Mixture/Distributed).
\\

Mediation/rest.
&
API paths
$0.749\pm0.089$ / $0.683\pm0.074$;
matched nulls
$0.150\pm0.017$ / $0.121\pm0.022$.
\\

Selectivity/graph
&
7/64 dimensions survive BH-FDR;
graph sparsity $0.791\pm0.023$ and
cross-seed Jaccard $0.83$.
\\

\bottomrule
\end{tabular}
\vspace{-1mm}
\end{minipage}
\end{center}

Table~\ref{tab:mechanistic_summary} summarizes three independent properties of the recovered interface. In the main figure, Shared is highest on target correctness, locality, copy preservation, and the resulting composite. The same architecture ordering persists for same-family, cross-family, and matched-random donors. The reduction under cross-family donors limits the claim to partially aligned causal roles, while the much lower matched-random condition shows that intervention scale and layer compatibility alone do not explain the result. The supplementary mediation analysis provides an independent criterion: API-aligned paths mediate 0.749 of the target effect compared with 0.150 for matched null paths, and necessity--restoration produces a similar separation.

Figure~\ref{fig:transplant} explains these aggregate results. Panel A inserts only the donor interface code while retaining the surrounding base computation. Donor and base episodes are matched at the same producer role, insertion position, difficulty range, output length, and intervention norm. Panel B maps the recovered producer--consumer routes and distinguishes family-specific APIs from a shared cross-channel. Panel C uses a common higher-is-better orientation and shows that Shared leads all four architectures on TC, LOC, CopyPres, and CompScore. The joint criterion prevents an architecture from appearing successful merely because it changes the target output while damaging unrelated state or copying donor-specific content.

Panel D separates three donor regimes. Same-family donors preserve both producer role and family-level semantics; cross-family donors preserve insertion position and intervention scale while changing the semantic family; matched-random donors preserve the intervention load without preserving causal identity. The monotonic reduction from same-family to cross-family and matched-random transfer therefore supports role-aligned reuse rather than unrestricted exchangeability. The remaining cross-family effect is interpreted conservatively: it may arise from partial structural overlap between families, generic compatibility of the receiving layer, or shared low-level computational features, and does not establish a universal communication bus.

Panel E shows that causal effects concentrate on a sparse subset of role-aligned paths. The graph has sparsity $0.791\pm0.023$ and cross-seed Jaccard overlap 0.83, indicating that the reported structure is not produced by one unstable seed or one dense patchability map. Path-blocked mediation, necessity--restoration, and matched-null comparisons further distinguish naturally used routes from directions that are merely intervention-accessible. Supplementary operation matrices, raw donor-copy failures, graph-threshold sweeps, and path-matching controls are reported in \suppfig{4}--\suppfig{5} and \supptab{4}, \supptab{10}.

\paragraph{Statistical protocol.}
All detailed comparisons preserve the paired experimental structure:
candidate interfaces within one model use identical episode splits,
future branches, interface seeds, and prompt-generation seeds.
Architecture differences are computed within matched
model--family--seed blocks and aggregated across families.
Sharedness-Gain and FSD confidence intervals use paired
family-level bootstrap resampling; permutation tests preserve model,
family, and seed assignments. Dimension- and path-level discoveries
use Benjamini--Hochberg FDR. Layer, rank, graph threshold, and fidelity
margin are selected on validation data and frozen before test
evaluation.

The representation-identity analysis asks whether the recovered effect
can be explained by an arbitrary low-dimensional or output-aligned
subspace. \suppfig{7} places the recovered API and J-space in the same
high-performing region, with no significant difference under the
stated TOST criterion. The supported conclusion is therefore practical
equivalence within the declared margin, rather than strict dominance
over J-space. SAE/Transcoder, PCA, output-gradient, answer-token, and
random subspaces remain weaker on at least one of future fidelity,
causal transplantation, locality, or matched-capacity necessity
(\supptab{9}). Rank- and sparsity-matched controls, selective ablation,
and load-matched recovery further show that the result is not explained
by dimensionality alone, while also revealing finite causal capacity.

\section{Discussion}

Forked futures define hidden-state equivalence through distributions
over operations issued only after the state has formed. Under this
criterion, Shared achieves a shorter prequential causal description
than Local, Mixture, and Distributed while retaining comparable mean,
worst-family, and tail FSD. The result is therefore an economy-of-reuse
claim rather than a large predictive-fidelity gain.

The intervention evidence separates this coding advantage from generic
decodability or patchability. Same-family transplantation changes the
target while preserving non-target state and limiting donor copying.
Cross-family and matched-random donors weaken the effect, bounding the
claim to partially aligned causal roles. Path blocking, restoration,
sparse graph structure, and matched nulls indicate that multiple
consumers use the recovered routes. Model organisms further distinguish
known routing organizations, although a false-Shared case limits claims
of perfect identifiability.

\section{Limitations and Scope}

The conclusions are limited to the tested models, structured task
families, selected layers, candidate architectures, and finite operation
bank. New consumers may refine or split the empirical quotient, while
natural-family holdout remains more controlled than open-domain dialogue,
tool use, multilingual interaction, or long-context reasoning. Synthetic
organisms may also be easier to recover than mechanisms formed during
large-scale pretraining.

Intervention conclusions depend on donor/base matching, locality
definitions, and the candidate path universe. Redundant routes can
compensate after blocking, while discrete locality may miss confidence
or reasoning-trajectory changes. Matched nulls, restoration, threshold
sweeps, and threshold-free recovery reduce but do not eliminate these
ambiguities (\suppfig{4}--\suppfig{5}; \supptab{10}).

\section{Conclusion}

Forked futures cast shared-state discovery as a falsifiable
architecture-selection problem. Shared provides a shorter held-out
causal description at comparable future-signature fidelity, while
transplantation, mediation, matched baselines, and model-organism tests
support selective reuse and expose its limits. Within this scope, the
models exhibit a compact reusable causal API.

\clearpage
\bibliographystyle{iclr2026_conference}
\bibliography{iclr2026_conference}

\clearpage
\appendix

\setcounter{figure}{0}
\setcounter{table}{0}
\setcounter{equation}{0}
\renewcommand{\thefigure}{S\arabic{figure}}
\renewcommand{\thetable}{S\arabic{table}}
\renewcommand{\theequation}{S\arabic{equation}}

\section{Supplement Overview}

\begin{figure*}[!tbp]
    \centering
    \includegraphics[width=\textwidth]{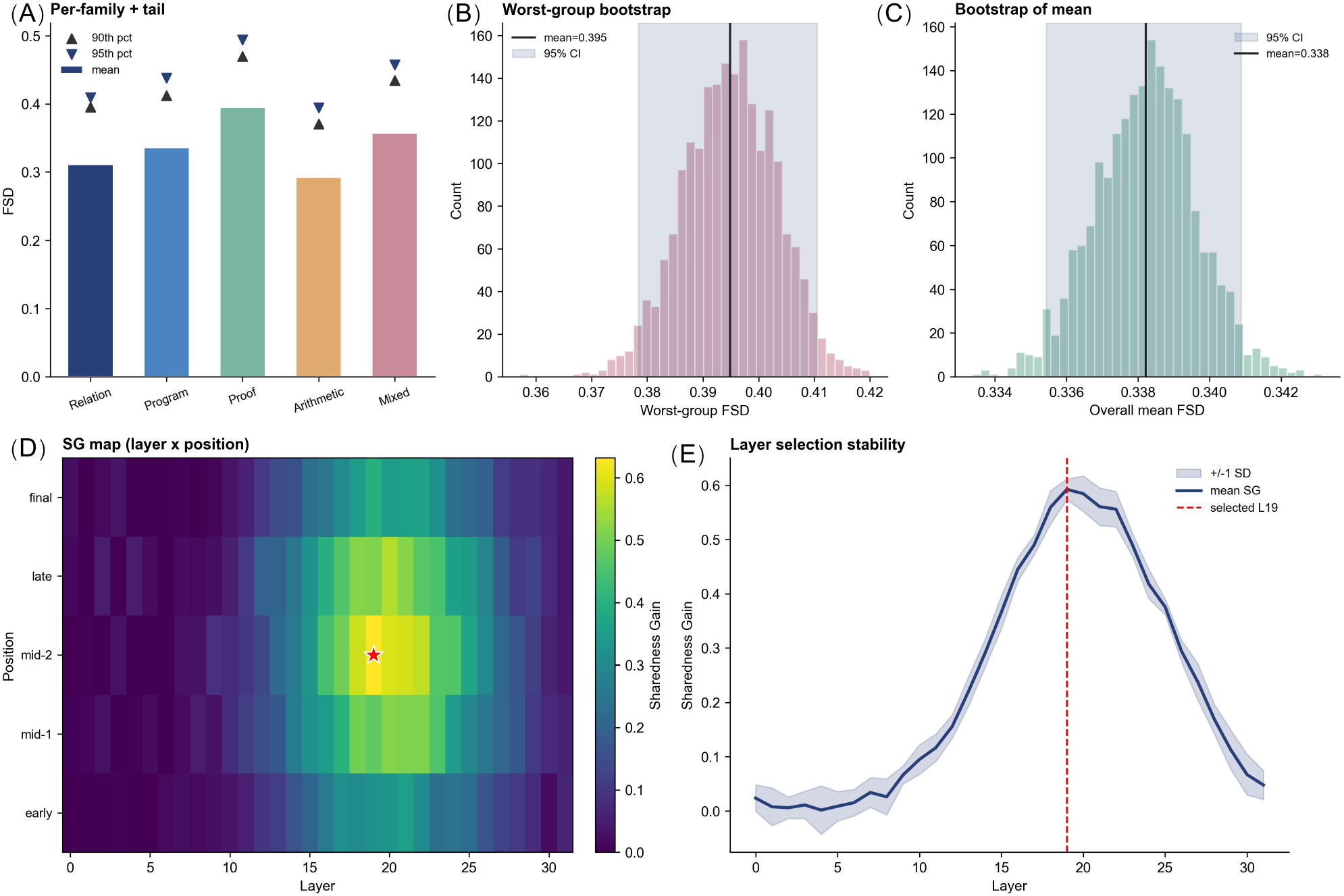}
    \caption{\textbf{Family-wise distortion, bootstrap uncertainty, and layer-position stability.} (A) Mean, 90th-percentile, and 95th-percentile FSD by family. (B) Bootstrap distribution of worst-group FSD. (C) Bootstrap distribution of overall mean FSD. (D) Sharedness-Gain heat map over layer and token position, with the validation-selected configuration marked. (E) Cross-seed mean Sharedness Gain by layer with a $\pm1$ standard-deviation band and selected layer.}
    \label{fig:slayers}
\end{figure*}

\begin{figure*}[!tbp]
    \centering
    \includegraphics[width=\textwidth]{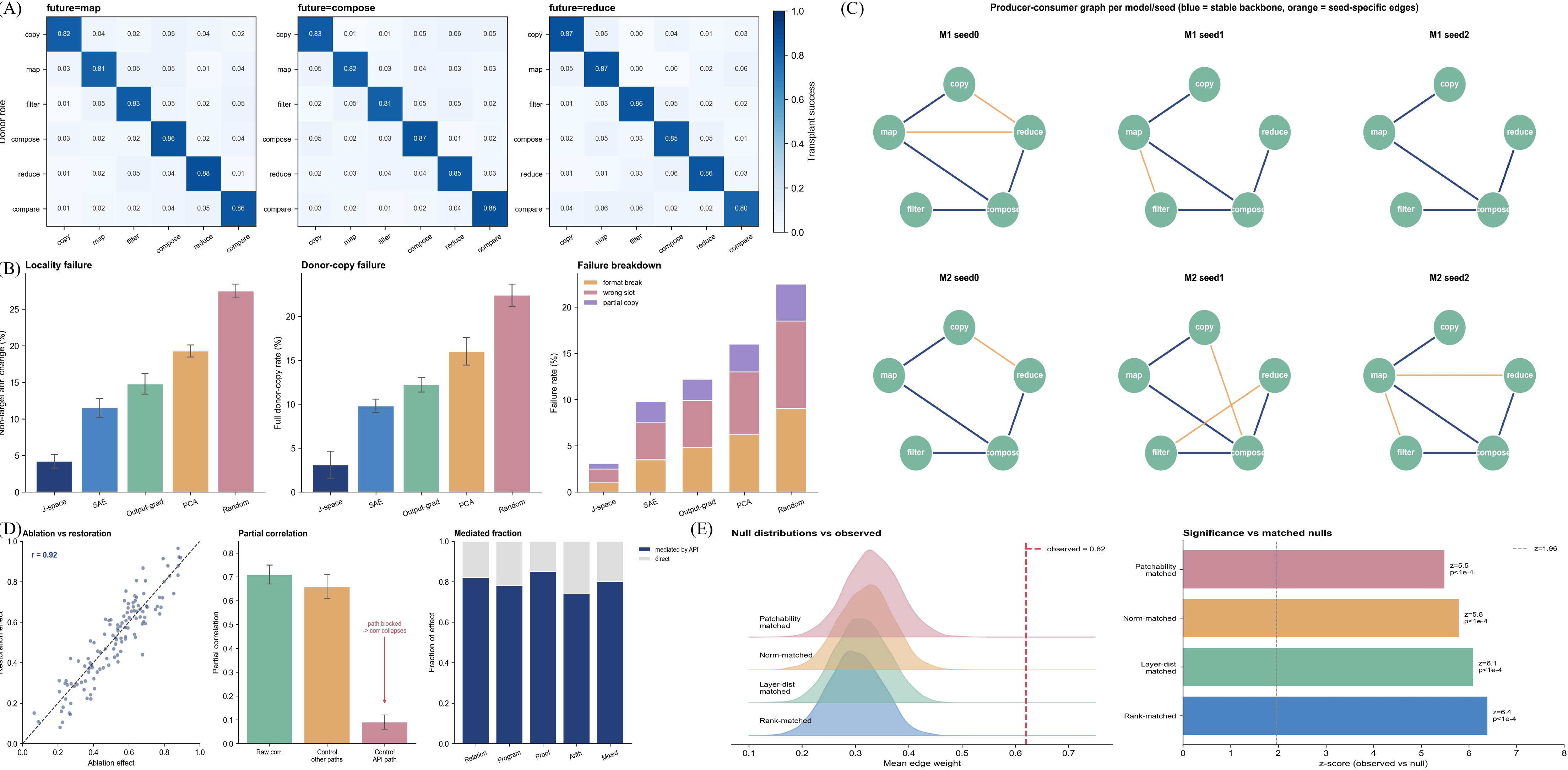}
    \caption{\textbf{Operation-specific transplantation, failure decomposition, graph stability, and causal mediation.} (A) Donor-role by consumer-role transplantation matrices for map, compose, and reduce futures. (B) Locality failure, donor-copy failure, and failure-type decomposition for the recovered interface and matched baselines. (C) Producer--consumer graphs across two models and three seeds; blue edges are the stable backbone and orange edges are seed-specific. (D) Ablation versus restoration, partial correlations before and after API-path blocking, and mediated fraction by family. (E) Observed edge strength against rank-, layer-distance-, norm-, and patchability-matched nulls.}
    \label{fig:smechanism}
\end{figure*}

This supplement expands the data construction, formal definitions, architecture competition, intervention protocol, falsification tests, robustness analyses, and reproducibility record used in the main paper. Figures~\ref{fig:sdata}--\ref{fig:srobustness} form five complementary evidence blocks: data validity, optimization and signature sensitivity, family/layer stability, mechanistic intervention, and robustness under threshold or representation changes. Figures~\ref{fig:soverview}--\ref{fig:sidentity} additionally preserve the conceptual overview and representation-identity analysis moved from the main paper. Tables~\ref{tab:scope}--\ref{tab:training_compute} provide the numerical, statistical, and reproducibility record supporting the reported claims.

Each figure and table is followed by a dedicated interpretation paragraph. These paragraphs state what the result directly establishes, identify the relevant controls, and distinguish the supported conclusion from claims that remain outside the tested scope. All architecture selection is performed on validation data; test splits are used only for the reported comparisons.

This long-form supplement provides a dedicated multi-paragraph section for every supplementary figure and table. The sections integrate panel- or row-level numerical interpretation, the corresponding causal argument, alternative explanations, and the precise boundary of the conclusion. No bullet-style analysis labels are used. Figures and tables are allowed to float to adjacent pages to keep the two-column layout filled; the text always refers to the numbered object explicitly.

\section{Data Construction and Controls}

\subsection{Forkable Episodes}

Each episode begins with a latent structured state $z$, a renderer $g$, and an operation bank $\mathcal K$.
The renderer produces a prefix $x_{1:t}=g(z)$ without revealing which future operation will be queried.
After the hidden state $h_{\ell,t}$ is cached, an operation $k$ is sampled and rendered as a continuation.
This ordering prevents the interface encoder from receiving future-operation text through the cached prefix.
The supervision package contains the current answer, the response distribution for each sampled future, generator metadata, and split tags.

\begin{figure*}[!tbp]
    \centering
    \includegraphics[width=\textwidth]{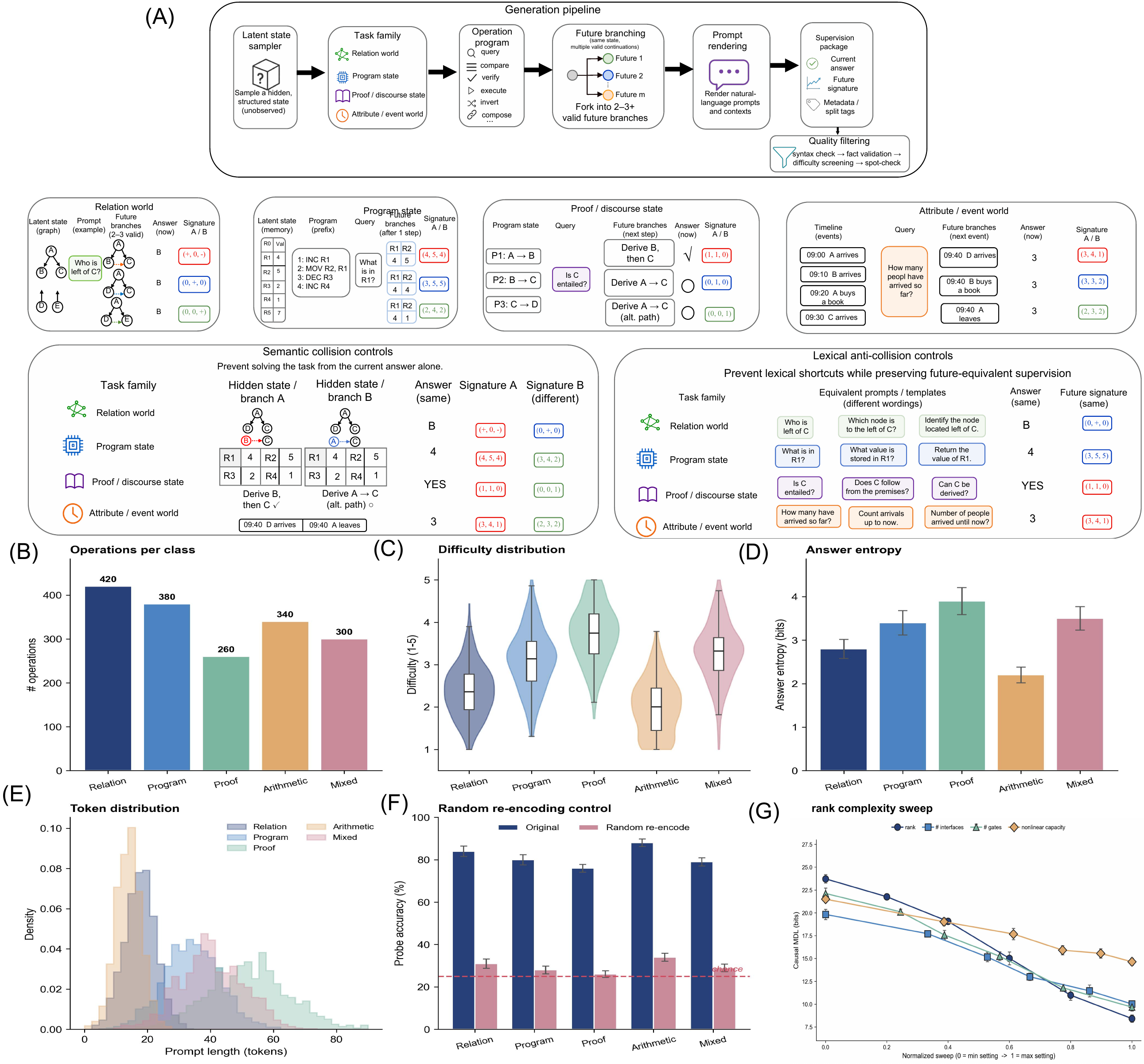}
    \caption{\textbf{Data construction, collision controls, and dataset diagnostics.} (A) The generation pipeline samples a latent structured state, selects one of four task families, instantiates an operation program, forks the same state into multiple valid futures, renders prompts, and packages current-answer, future-signature, metadata, and split supervision after quality filtering. The lower half gives concrete family examples, semantic-collision pairs that share the current answer but differ in future signature, and lexical anti-collision groups that vary wording while preserving future identity. (B) Number of operations by class. (C) Difficulty distributions. (D) Answer entropy. (E) Prompt-length distributions. (F) Probe performance before and after random symbol re-encoding. (G) Rank/complexity sweep.}
    \label{fig:sdata}
\end{figure*}

We use four generation families in the extended suite:
The \textbf{relation-world} family represents a latent directed graph and supports queries, comparisons, inversion, and multi-hop composition.
The \textbf{program-state} family represents register-like memory and supports execution, verification, copying, increment/decrement operations, and multi-step updates.
The \textbf{proof/discourse-state} family represents premise and rule graphs and supports entailment, alternate derivations, and chained reasoning.
The \textbf{attribute/event-world} family represents structured timelines and supports state updates, counting, temporal filtering, and event composition.
The main architecture comparisons use the first three natural families so that leave-one-family-out evaluation has three folds.
The attribute/event family is used in the extended construction and robustness analyses in Fig.~\ref{fig:sdata}.

\subsection{Semantic and Lexical Collision Controls}

A semantic collision pair has the same current answer but different future signatures.
For example, two program states may both return the same current register value while diverging after an increment or move operation.
Such pairs make current-answer shortcuts insufficient.
A lexical anti-collision group expresses the same latent query with multiple templates while preserving its future signature.
Random entity re-encoding and novel generator templates are assigned after split construction.
The controls therefore separate three quantities: current-answer identity, future causal identity, and surface-form identity.

\subsection{Supplementary Figure S1: Data Validity, Collision Controls, and Benchmark Heterogeneity}

The operation counts shown in panels B--E are heterogeneous rather than balanced by construction: relation contains approximately 420 operations, program 380, proof/discourse 260, attribute/event 340, and the mixed condition about 300. Proof/discourse has the smallest operation bank but the highest median difficulty and answer entropy, close to four bits, whereas attribute/event examples are shorter and closer to two bits of entropy. Prompt lengths also differ markedly, with proof/discourse producing the longest upper tail. The interface is therefore evaluated across families that differ simultaneously in uncertainty, context length, and compositional depth. The semantic-collision examples preserve the present answer while changing the future response vector, so a current-answer probe cannot solve the benchmark. The lexical anti-collision examples impose the complementary requirement: multiple surface forms share the same latent state and future signature. Panel F strengthens this conclusion because random symbol re-encoding reduces surface-form probes from roughly 76--88\% to approximately 25--34\%, close to the lexical-control floor. The result is consequently tied to future state structure rather than stable entity names or one prompt template. Panel G shows that Shared remains favored across a broad rank/complexity region, but the exact numeric gain changes with the coding penalty.

This supports robustness to reasonable capacity choices, not invariance to every possible nonlinear encoder or every future operation bank. A substantially richer bank may refine the causal quotient and change the preferred rank. The construction also provides a useful diagnostic for whether the future signature is rich enough. If all semantic-collision pairs had nearly identical signatures, the benchmark would collapse back to answer supervision even though the prompts were formally forked. Conversely, if lexical anti-collision groups produced large signature differences, the measured quotient would be contaminated by wording. The examples in panel A and the re-encoding results in panel F jointly test both directions: states that should differ remain separable after the answer is matched, and states that should agree remain close after the wording is changed. This paired design is stronger than reporting average probe accuracy because it explicitly identifies the equivalence and non-equivalence relations that the interface is expected to preserve. The quality-control pipeline at the bottom of panel A is equally important.

Syntax validation removes malformed prompts, fact validation checks that each continuation is licensed by the latent state, difficulty screening prevents trivial one-token shortcuts from dominating, and manual spot checks inspect generator failure modes that are difficult to encode automatically. These filters define the empirical causal object before any interface is trained. If invalid continuations entered the signature cache, a low-FSD encoder could be rewarded for reproducing generator errors rather than model behavior. The supplement should therefore retain the generator version, validation scripts, rejection counts, and examples of failed episodes in the released artifact. Finally, the family imbalance should be handled at the level of statistical units rather than by forcing equal example counts. Proof/discourse has fewer operations but higher entropy and longer prompts; oversampling it until the raw counts match could distort the natural difficulty distribution. The paper instead aggregates at the task-family level for confidence intervals and reports family-specific distortion. This preserves the harder family as a meaningful stress test while preventing the largest relation or program pool from dominating significance merely through sample size.

\begin{table*}[!tbp]
\centering
\caption{\textbf{Experimental scope, split hierarchy, and falsification coverage.} The table records model scale, task families, operation banks, held-out operation/composition/generator/family splits, seed structure, model-organism configuration, competing architectures, signature definitions, and pre-specified tests.}
\label{tab:scope}
\small
\begin{tabular}{p{0.29\textwidth}p{0.66\textwidth}}
\toprule
Dimension & Configuration \\
\midrule
Detailed models & Qwen2.5-1.5B ($d=1536$, 28 layers); Llama-3-8B ($d=4096$, 32 layers) \\
Directional SG sweep & Qwen2.5-7B, Mistral-7B, Gemma-2-9B, and Phi-3-3.8B; architecture-family, intervention, and mediation claims remain based on the two detailed models \\
Main task families & Relation world, Program state, Proof/discourse state \\
Training operations per family & 16 operations spanning query, compare, verify, execute, invert, and compose \\
Held-out operations & 4 per family, unseen during interface training \\
Held-out compositions & Two-step and three-step chained operations \\
Held-out generators & Template variants with novel entity and relation sets \\
Natural-family holdout & Leave-one-family-out cross-validation (three folds) \\
Interface initialization seeds & 5 per model and configuration \\
Prompt-generation seeds & 5 per interface seed \\
Model-organism hard-label audit & 4 held-out organisms per class ($16$ total; $d=256$, 6 layers, vocabulary 512, 3000 steps) \\
Competing architectures & Shared, Local, Mixture, Distributed \\
Signature definitions & Full-vocabulary logits, top-100, answer-token, sequence log-likelihood \\
Statistical tests & Bootstrap confidence intervals, permutation tests, mixed-effects analysis, TOST, and BH-FDR \\
\bottomrule
\end{tabular}
\end{table*}

\subsection{Supplementary Table S1: Scope, Split Hierarchy, and Falsification Coverage}

The evaluation crosses five interface initializations with five prompt-generation seeds, producing twenty-five model/data configurations per detailed model before aggregation. Four held-out operations are reserved per family, and the hierarchy also includes unseen two- and three-step compositions, unseen generators, and three leave-one-family-out folds. The hard-label model-organism audit uses four held-out organisms for each of the four known architecture classes; Supplementary Fig.~S2 additionally visualizes posterior diagnostics over an extended synthetic calibration sweep. Each split removes a different shortcut. Operation holdout tests whether an existing state supports a new consumer; composition holdout tests systematic reuse of known primitives; generator holdout changes the rendering process; family holdout changes the latent state semantics. Agreement across the hierarchy is stronger than success on a single random train/test division. The separation of interface seeds from prompt-generation seeds also prevents optimization variability from being confused with data-rendering variability. The table defines a broad controlled evaluation, but all families still arise from structured generators.

Leave-one-family-out testing is therefore stronger than template OOD but weaker than arbitrary natural conversation, open-ended tool use, or multilingual interactive settings. The scope table also clarifies which choices are fixed before test evaluation. The operation inventory, family-wise fidelity thresholds, TOST margin, layer-selection rule, and prequential block order must be decided using training or validation data. If any of these are tuned after observing the test MDL, the architecture competition becomes vulnerable to researcher degrees of freedom. The released configuration should therefore include a machine-readable manifest that maps every sample to its family, generator, operation, composition depth, language condition, and prequential block, together with the seed that produced it. A second benefit of the table is that it separates replication from robustness. Qwen2.5-1.5B and Llama-3-8B constitute cross-family replication of the architecture effect, while alternative signatures, layers, languages, and thresholds are robustness analyses within that broader claim.

These should not be pooled into one undifferentiated average. A reviewer should be able to ask whether the effect is present in the worst model, the worst family, the hardest split, and the least favorable reasonable hyperparameter setting. The supplement provides those strata rather than reporting only a grand mean. The model-organism row defines a distinct falsification experiment rather than another benchmark. Because the ground-truth routing structure is known, architecture accuracy, false-Shared rate, MDL margin, and path overlap can be evaluated directly. This is the only setting in which the paper can say that the discovery procedure recovered the correct architecture rather than merely that Shared fit real-model data better than the chosen alternatives. Keeping that distinction explicit prevents synthetic ground truth from being used to overclaim identifiability in pretrained models.

\section{Formal Definitions}

\subsection{Future Causal Signature}

For model $p_\theta$, prefix state $h$, and future operation $k$, define
\begin{equation}
 \sigma_k(h)=p_\theta(y\mid h,k).
\end{equation}
The empirical signature over sampled operations $\mathcal K_h$ is
\begin{equation}
 \widehat{\Sigma}_{\mathcal K}(h)
 =\{(k,\sigma_k(h)):k\in\mathcal K_h\}.
\end{equation}
For full-vocabulary signatures, the distortion between an original and reconstructed state is
\begin{equation}
 \FSD(h,\hat h)
 =\frac{1}{|\mathcal K_h|}
  \sum_{k\in\mathcal K_h}
  \JS\!\left(\sigma_k(h)\,\|\,\sigma_k(\hat h)\right).
\end{equation}
Top-$K$ signatures renormalize the union of the two top-$K$ supports.
Answer-token signatures restrict the distribution to task-defined answer tokens.
Sequence-LL signatures compare normalized log-likelihood vectors over valid continuation sequences.

\subsection{Architecture Classes}

Let $r$ be the interface rank and $m$ the number of operations.
The Shared model uses $c=E(h)\in\mathbb R^r$ and $\hat h=R(c)$.
The Local model uses $c_k=E_k(h)$ and $\hat h=R_k(c_k)$.
The Mixture model uses candidate codes $\{c_j\}_{j=1}^J$ with a gate $q(j\mid h,k)$.
The Distributed model reconstructs from multiple components without a privileged bottleneck.
For matched-rank comparisons, every active code presented to a consumer has dimension $r$.
For matched-parameter comparisons, widths are adjusted so that total trainable parameters agree with the Shared or Local budget.

\subsection{Prequential Causal MDL}

The dataset is ordered into blocks $\mathcal D_1,\ldots,\mathcal D_B$ before architecture training.
An uninformed code transmits the first block.
For $b>1$, the interface is fitted on $\mathcal D_{<b}$ and evaluated on $\mathcal D_b$:
\begin{equation}
 \MDL_{\mathrm{causal}}(A)
 = L_0(\mathcal D_1)
 +\sum_{b=2}^B
 \sum_{i\in\mathcal D_b}
 -\log p_{\hat\eta_{A,<b}}(\Sigma_i\mid h_i,k_i).
\end{equation}
The same block order is used for all architectures within a seed.
Sharedness Gain is computed from held-out blocks:
\begin{equation}
 \SG =
 \min\{\MDL_{\mathrm{Local}},
       \MDL_{\mathrm{Mixture}},
       \MDL_{\mathrm{Distributed}}\}
 -\MDL_{\mathrm{Shared}}.
\end{equation}
We require both $\SG>0$ and satisfaction of the pre-specified family-wise fidelity constraint.

\section{Training Diagnostics and Signature Sensitivity}

\begin{figure*}[!tbp]
    \centering
    \includegraphics[width=\textwidth]{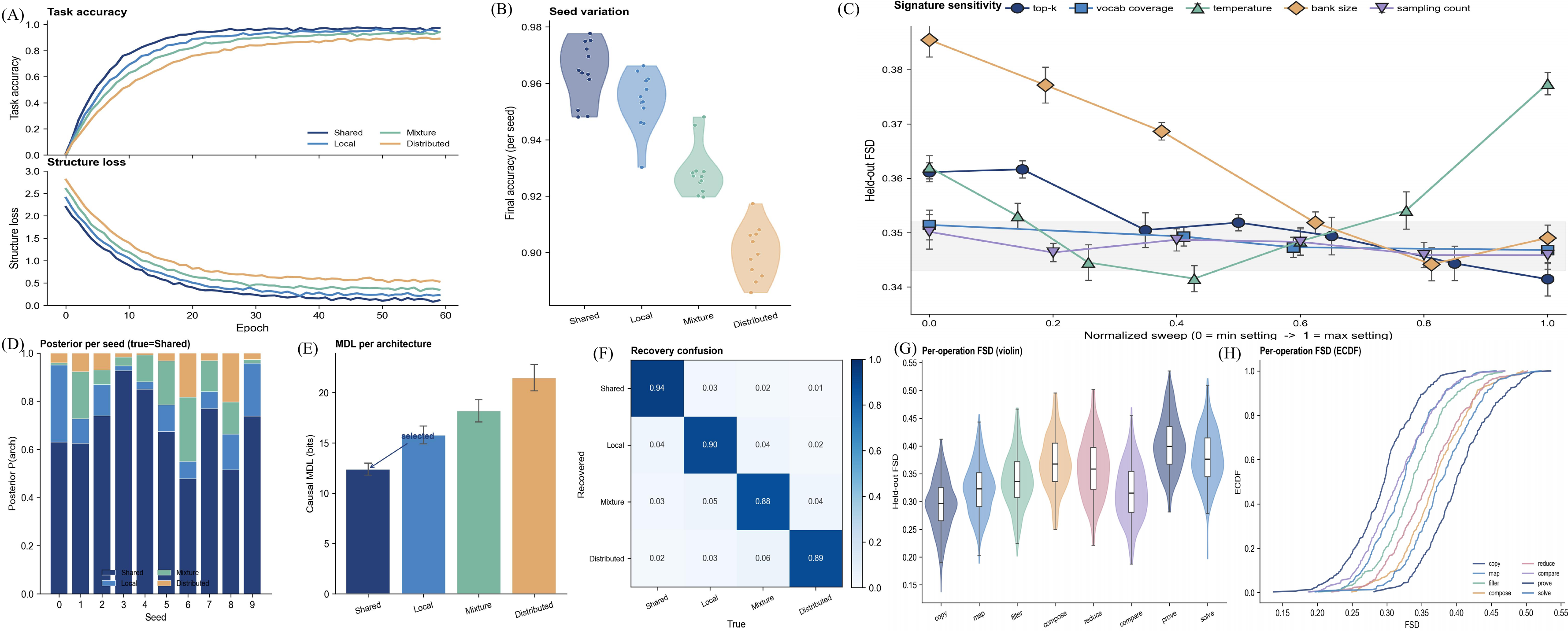}
    \caption{\textbf{Optimization, seed variation, signature sensitivity, and blind architecture recovery.} (A) Task-accuracy and structure-loss trajectories for all architectures. (B) Final per-seed accuracy distributions. (C) Held-out FSD under top-$k$, vocabulary-coverage, temperature, bank-size, and sampling-count sweeps. (D) Posterior architecture mass for true-Shared organisms. (E) Causal MDL by architecture. (F) Normalized posterior recovery matrix from the extended synthetic diagnostic sweep; hard-label held-out counts are reported in Table~\ref{tab:organisms} and the main figure. (G) Per-operation FSD distributions. (H) Empirical cumulative distributions of operation-level FSD.}
    \label{fig:straining}
\end{figure*}

Figure~\ref{fig:straining} reports optimization trajectories, seed variation, signature alternatives, posterior model-organism recovery, per-architecture MDL, class confusion, and operation-level FSD distributions.
The agreement among full-vocabulary, top-$100$, answer-token, and sequence-LL signatures indicates that the main architecture ordering is not tied to one output summary.

\subsection{Supplementary Figure S2: Optimization, Signature Sensitivity, and Architecture Recovery}

By the end of training, Shared reaches task accuracy near 0.97 and structure loss near 0.2; Local is approximately 0.95/0.3, Mixture 0.93/0.45, and Distributed 0.90/0.6. Final-seed distributions preserve this ordering. In the signature sweep, moderate top-$k$, vocabulary-coverage, temperature, operation-bank, and sampling settings keep FSD around 0.345--0.352, whereas severe bank undercoverage or extreme temperature raises FSD toward 0.38. The architecture comparison is not being driven by an obviously failed optimizer: all candidates learn the underlying task, but Shared reaches the best accuracy-loss trade-off and the shortest prequential code. The model-organism posterior and MDL panels show concentrated but non-degenerate class evidence in the extended synthetic diagnostic sweep. The normalized matrix remains broadly diagonal, while the separate hard-label audit in Table~\ref{tab:organisms} records 14/16 correct decisions and one false-Shared error. Operation-level violins and ECDFs show broader and higher-tailed distortion for proof, comparison, and some transform operations. A stable pooled mean therefore does not imply uniform performance.

This motivates reporting per-operation, worst-family, and tail FSD rather than describing one global average as complete sufficiency. The optimization curves should also be read alongside the prequential protocol. The relevant quantity is not the final training loss of one model trained on all blocks, but the cumulative code produced by a sequence of models that only see earlier data. A candidate can have a competitive final loss and still incur a long description if it requires many blocks before generalizing. Shared converges quickly enough that its advantage appears throughout the sequence rather than only after saturation. This is precisely the behavior expected from a reusable representation: information learned for one operation or family should reduce the code length of later related futures. Panel C provides a practical criterion for selecting the signature operating region. The shaded band is not merely where FSD is numerically lowest; it is where the ranking and tail behavior remain stable while the signature retains sufficient probability mass.

Very small top-$k$ or bank-size settings underrepresent the output distribution, while excessive temperature flattens meaningful distinctions. The supplement should report the exact chosen point and demonstrate that it lies inside the stable region rather than at its boundary. The model-organism panels reveal a subtle limitation. Posterior mass for Shared is dominant but not always close to one, indicating that finite data can leave Local as a plausible runner-up even when the true architecture is shared. This motivates reporting uncertainty over architecture classes and the MDL margin, not only the argmin label. In real models, where the ground truth may be partially shared or redundant, close posterior mass should be interpreted as ambiguity rather than forced into a categorical conclusion.

\subsection{Supplementary Figure S3: Family Tails, Bootstrap Uncertainty, and Layer Stability}

Mean FSD is approximately 0.31 for relation, 0.335 for program, 0.395 for proof, 0.29 for arithmetic/event, and 0.355 for mixed examples. The 90th- and 95th-percentile markers lie roughly 0.07--0.12 above the means. The worst-group bootstrap is centered near 0.395, while the bootstrap of the pooled mean is centered near 0.338 and is substantially narrower. Panels D--E show that Sharedness Gain is not a single-layer spike. It begins increasing around layers 10--12, forms a broad high-value region from approximately layers 16--23, and peaks around layer 19 at the mid-state position. The independent layer-selection curve peaks in the same region with a comparatively narrow cross-seed band. The result identifies a stable stage at which the tested operation families share future-relevant state. It does not imply that early or final layers contain no relevant information, nor that layer 19 is a universal API layer for other architectures or tasks. The family and layer diagnostics also constrain causal interpretation. The high proof/discourse tail could arise from longer contexts, more uncertain valid continuations, or a genuine mismatch between the selected rank and proof-state complexity. The current figure cannot fully distinguish these sources, so the supplement should retain matched-length and matched-entropy controls where available.

What it does establish is that the Shared result is not obtained by ignoring the hardest family: the selection rule explicitly monitors the family whose tail is largest. The layer-position map suggests a producer--consumer chronology. Early positions and layers have low SG because the latent state has not yet been integrated into a compact future-relevant code. The gain rises at middle positions after the structured prefix has been processed, then decreases near the final output where operation-specific decoding becomes dominant. This interpretation is consistent with, but not proven solely by, the heat map. It gains credibility from the later path-mediation results, which place producer hubs in the middle layers and consumer sinks in later layers. The independent layer-selection curve is designed to avoid circularity. The layer is chosen from validation seeds and then frozen for the reported test comparisons. Reporting the full curve ensures that a reviewer can see whether the chosen layer lies on a broad plateau or a narrow noise peak. A useful additional release artifact is the per-seed curve before averaging, because a stable mean can conceal seeds whose maxima occur in qualitatively different regions.

\begin{table*}[!tbp]
\centering
\caption{\textbf{Architecture competition and future-signature sufficiency on two model families.} MDL is held-out prequential causal description length; mean, worst-family, and 95th-percentile FSD quantify fidelity; parameter counts expose capacity differences; SG compares Shared with the best non-Shared alternative. Confidence intervals use 10,000 paired bootstrap resamples.}
\label{tab:architecture}
\small
\resizebox{\textwidth}{!}{
\begin{tabular}{llrrrrrr}
\toprule
Model & Architecture & MDL$\downarrow$ & Mean FSD$\downarrow$ & Worst family & Tail$_{95}$ & Parameters & SG [95\% CI] \\
\midrule
\multirow{4}{*}{Qwen2.5-1.5B}
& Shared & 1.292 & 0.341 & 0.358 & 0.372 & 98K & \multirow{4}{*}{$+0.216\ [0.161,0.278]$}\\
& Local & 1.508 & 0.343 & 0.361 & 0.379 & 98K & \\
& Mixture & 1.672 & 0.349 & 0.367 & 0.385 & 103K & \\
& Distributed & 1.981 & 0.352 & 0.371 & 0.392 & 2.5M & \\
\midrule
\multirow{4}{*}{Llama-3-8B}
& Shared & 1.187 & 0.328 & 0.344 & 0.361 & 262K & \multirow{4}{*}{$+0.294\ [0.208,0.391]$}\\
& Local & 1.481 & 0.331 & 0.349 & 0.368 & 262K & \\
& Mixture & 1.594 & 0.336 & 0.358 & 0.374 & 270K & \\
& Distributed & 1.863 & 0.339 & 0.362 & 0.381 & 8.4M & \\
\bottomrule
\end{tabular}}
\par\smallskip
\footnotesize\textit{Note:}
Shared has the lowest MDL on both model families while maintaining comparable or lower FSD.
The separation is larger on Llama-3-8B (SG $=0.294$).
Permutation tests give $p<0.001$ for both models.
\end{table*}

\subsection{Supplementary Table S2: Architecture Competition on Two Model Families}

The complete architecture, family-stratified, intervention, and uncertainty analyses use Qwen2.5-1.5B and Llama-3-8B. The main-paper Sharedness-Gain panel additionally includes Qwen2.5-7B, Mistral-7B, Gemma-2-9B, and Phi-3-3.8B as a directional robustness sweep; those four backbones are not treated as full replications of the intervention suite.

For Qwen2.5-1.5B, Shared has MDL 1.292 and Local 1.508, giving SG $=0.216$; the gaps to Mixture and Distributed are 0.380 and 0.689 nats. For Llama-3-8B, Shared is 1.187 and Local 1.481, giving SG $=0.294$; the gaps to Mixture and Distributed are 0.407 and 0.676. The task-family bootstrap intervals, $[0.161,0.278]$ and $[0.208,0.391]$, exclude zero. Shared and Local have the same approximate parameter budget on each model---98K on Qwen and 262K on Llama---yet Local has a longer description. The result is therefore not a trivial parameter-count comparison. Mean FSD differs by only 0.011 or less within a model, but Shared is also best on worst-family and 95th-percentile distortion. The appropriate conclusion is shorter causal description at equivalent or slightly better fidelity. The table rejects the tested Local, Mixture, and Distributed organizations under the stated coding contract. It does not prove uniqueness against every possible nonlinear, redundant, or hierarchically shared architecture. The prequential MDL values can be decomposed conceptually into how quickly the interface learns reusable structure and how much architecture-specific information must be transmitted. Local pays repeatedly for operation-specific parameters or predictions, Mixture pays for both component structure and gating decisions, and Distributed pays for maintaining several active components.

Shared avoids those repeated costs when later operations reuse distinctions already learned from earlier ones. The table's importance is therefore not only that Shared has fewer nats, but that the savings occur under held-out operations where memorizing the training operation identity cannot help. The effect size should be interpreted relative to the bootstrap unit. Resampling individual prompts would yield artificially narrow intervals because prompts from the same family and generator share structure. Task-family-level paired bootstrap preserves the dependence that matters for the scientific claim. The intervals remain positive even under this conservative unit, which is stronger evidence than an example-level confidence interval with a much larger effective sample size. At the same time, the small absolute FSD differences require disciplined wording. The result does not show that Shared preserves dramatically more future information. It shows that the same fidelity can be achieved with a shorter causal description and that Shared is non-inferior in the worst group. This distinction should appear consistently in the abstract, results, and captions. Overstating the result as a large predictive improvement would invite the correct criticism that the architecture ranking is driven mainly by complexity control.

\begin{table*}[!tbp]
\centering
\caption{\textbf{Architecture ranking across future-signature definitions.} Rows compare full-vocabulary, top-100, answer-token, and sequence log-likelihood signatures; the final rows summarize top-$k$ and temperature sensitivity.}
\label{tab:signature}
\small
\resizebox{\textwidth}{!}{
\begin{tabular}{lrrrrr}
\toprule
Signature & Shared & Local & Mixture & Distributed & SG \\
\midrule
Full vocabulary & 1.292 & 1.508 & 1.672 & 1.981 & 0.216 \\
Top-100 tokens & 1.318 & 1.527 & 1.695 & 2.014 & 0.209 \\
Answer token & 0.847 & 1.031 & 1.162 & 1.389 & 0.184 \\
Sequence log-likelihood & 1.104 & 1.342 & 1.518 & 1.791 & 0.238 \\
\midrule
\multicolumn{5}{l}{Sensitivity to top-$k$, $k\in\{50,100,200,500,\mathrm{full}\}$} & std $=0.018$ \\
\multicolumn{5}{l}{Sensitivity to temperature, $\tau\in\{0.5,1.0,1.5,2.0\}$} & std $=0.024$ \\
\bottomrule
\end{tabular}}
\end{table*}

\subsection{Supplementary Table S3: Dependence on Future-Signature Definition}

Shared is selected under all four output summaries: SG is 0.216 for full vocabulary, 0.209 for top-100, 0.184 for answer tokens, and 0.238 for sequence log-likelihood. The narrower answer-token signature yields the smallest separation, while the sequence-level signature gives the largest. Across $k\in\{50,100,200,500,\mathrm{full}\}$, SG has standard deviation 0.018; across temperatures 0.5--2.0, the standard deviation is 0.024. The unchanged ordering shows that the discovery is not specific to concatenated top-100 logits. Richer signatures preserve more uncertainty and produce at least as strong a Shared advantage, whereas the answer-token representation discards information needed by verification, inversion, and composition. Signature choice changes the absolute MDL scale and the effect magnitude. The table supports robustness over the tested summaries, not equivalence between every possible sequence-level metric or decoding procedure. The signature comparison also reveals which future information contributes most strongly to architecture discrimination. Answer-token signatures retain only the probability assigned to task-defined outputs and therefore erase differences in alternative valid continuations, calibration, and sequence structure. Their positive but smaller SG shows that even a narrow target contains some reusable state, but the stronger sequence and full-vocabulary results indicate that the hidden API is expressed in a richer distributional object.

The sensitivity rows should be read as a robustness envelope rather than an invitation to select the largest SG. The full-vocabulary signature is the default because it makes the weakest truncation assumption, not because one sweep point maximizes the effect. The reported standard deviations show how much the conclusion changes under reasonable preprocessing. A pre-specified analysis should specify the default before evaluation and use the sweep only to test whether the conclusion reverses. A remaining open question is whether two signatures that produce the same architecture ranking identify the same coordinates. The current table compares coding outcomes, not principal angles or dimension correspondences between signature-trained interfaces. The re-encoding and representation-identity analyses partially address geometry, but a complete invariance claim would require aligning interfaces learned from each signature definition and testing cross-signature transplantation.

\section{Transplantation, Mediation, and Selectivity}

\subsection{Intervention Protocol}

For a base episode $b$ and donor episode $d$, the intervention is applied only after both episodes reach the same producer role.
The donor interface code replaces the base code while the base residual context remains fixed.
The target continuation is scored for the donor-side relational or operational result.
Locality is evaluated on non-target outputs from the base episode.
Donor-copy rate counts outputs that reproduce donor-specific surface content not required by the target relation.

Matched-random donors preserve task family, code norm, layer distance, and intervention position.
Cross-family donors preserve the intervention position but change the task family.
Local-API donors use the operation-specific interface selected under the Local architecture.
Each comparison uses the same base examples.

\subsection{Mediation and Selectivity}

Producer--consumer edges are retained when blocking the corresponding path reduces the target effect and restoration of the API code recovers it.
Matched null edges preserve producer--consumer layer distance and intervention load.
Threshold-free AUPRC is reported in addition to graph statistics.
Dimension-level effects are compared with random directions and corrected using Benjamini--Hochberg FDR at $q=0.05$.

\begin{table*}[!tbp]
\centering
\caption{\textbf{Figure-aligned transplantation, donor controls, mediation, and dimension selectivity.} Panel-C and donor-control values reproduce the rounded higher-is-better summaries in the main transplantation figure. CopyPres is $1-\mathrm{DonorCopy}$ and the displayed CompScore is the equal-weight mean of TC, LOC, and CopyPres, computed from unrounded measurements.}
\label{tab:transplant}
\small
\resizebox{\textwidth}{!}{
\begin{tabular}{lrrrr}
\toprule
Architecture & TC$\uparrow$ & LOC$\uparrow$ & CopyPres$\uparrow$ & CompScore$\uparrow$ \\
\midrule
Shared & 0.91 & 0.88 & 0.85 & 0.87 \\
Local & 0.72 & 0.75 & 0.66 & 0.70 \\
Mixture & 0.65 & 0.69 & 0.61 & 0.64 \\
Distributed & 0.48 & 0.52 & 0.42 & 0.46 \\
\bottomrule
\end{tabular}}

\medskip
\resizebox{\textwidth}{!}{
\begin{tabular}{lrrr}
\toprule
Architecture & Same-family CompScore & Cross-family CompScore & Matched-random CompScore \\
\midrule
Shared & 0.88 & 0.78 & 0.53 \\
Local & 0.71 & 0.58 & 0.39 \\
Mixture & 0.64 & 0.51 & 0.34 \\
Distributed & 0.47 & 0.35 & 0.24 \\
\bottomrule
\end{tabular}}

\medskip
\resizebox{\textwidth}{!}{
\begin{tabular}{lrrrr}
\toprule
Mediation test & API paths & Null paths & Difference & $p$-value \\
\midrule
Mean mediated fraction & $0.749\pm0.089$ & $0.150\pm0.017$ & 0.599 & $<0.001$ \\
Necessity--restoration & $0.683\pm0.074$ & $0.121\pm0.022$ & 0.562 & $<0.001$ \\
Path-blocked, multi-consumer & $0.714\pm0.096$ & $0.138\pm0.019$ & 0.576 & $<0.001$ \\
\midrule
$N$ producer--consumer pairs & \multicolumn{4}{c}{25 API paths and 25 null-matched paths} \\
\bottomrule
\end{tabular}}

\medskip
\resizebox{\textwidth}{!}{
\begin{tabular}{lrrrrr}
\toprule
Selectivity test & Mean effect & Maximum effect & Key dimension & Ratio & Inference \\
\midrule
Absolute effect & 1.981 & 2.904 & 42 & -- & 95\% CI $[1.742,2.231]$ \\
Random direction & 0.083 & 0.217 & -- & $23.9\times$ & $p<0.001$ \\
BH-FDR & \multicolumn{5}{l}{7 of 64 dimensions pass correction at $q=0.05$} \\
\bottomrule
\end{tabular}}
\end{table*}

\subsection{Supplementary Table S4: Transplantation, Donor Controls, Mediation, and Selectivity}

The first block reproduces the main-figure architecture profiles. Shared is highest on target correctness, locality, copy preservation, and the equal-weight composite. Because the constituent bars are rounded to two decimals while CompScore is computed from unrounded values, recomputing the mean from the printed bars can differ by one percentage point. The second block preserves the architecture ordering under same-family, cross-family, and matched-random donors. The reduction from same-family to cross-family and matched-random transfer bounds the claim to aligned causal roles rather than unrestricted content exchange.

The displayed donor-copy quantity is CopyPres, defined as $1-\mathrm{DonorCopy}$, so every metric in the main figure has a higher-is-better orientation. The corresponding raw donor-copy events, operation-level matrices, and failure decompositions remain reported in Supplementary Fig.~S4. The composite is
\begin{equation}
\mathrm{CompScore}
=
\frac{\mathrm{TC}+\mathrm{LOC}+\mathrm{CopyPres}}{3},
\end{equation}
and is computed from unrounded constituent measurements. This definition prevents a method from appearing strong solely through target change while making its relation to the plotted components transparent.

The mediation results provide an independent natural-use test. API-aligned paths mediate $0.749\pm0.089$ of the target effect versus $0.150\pm0.017$ for matched null paths; necessity--restoration and multi-consumer blocking show similar separations. Dimension selectivity is sparse: API-aligned effects are much larger than random directions, while only 7 of 64 dimensions survive BH-FDR. Together, the results support selective, path-dependent reuse without claiming unrestricted cross-family exchange or coordinate uniqueness.

\subsection{Supplementary Figure S4: Role Selectivity and Producer--Consumer Mediation}

Role-aligned diagonal transplantation values are approximately 0.81--0.88 for map, compose, and reduce, while most off-diagonal cells remain between 0.00 and 0.06. Non-target error is about 4\% for the recovered interface, 11--12\% for SAE, 15\% for output-gradient, 19\% for PCA, and 27--28\% for random directions. Donor-copy failure rises from roughly 3\% for the interface to more than 22\% for random directions. The stable blue backbone in the producer--consumer graphs recurs across both models and multiple seeds. Ablation and restoration effects correlate at approximately $r=0.92$ before path blocking, while the partial correlation falls to about 0.09 when the API route is removed. Family-wise mediated fractions remain roughly 0.74--0.85. These results connect role selectivity, graph stability, necessity, restoration, and mediation within one intervention analysis. Observed edge strength near 0.62 exceeds matched-null centers around 0.28--0.34, with standardized effects of roughly 5.5--6.4.

This rules out simple layer-distance, rank, norm, or patchability explanations. It still does not imply that the displayed graph contains every redundant route used by the full transformer. The off-diagonal structure in panel A is a direct test of compositional specificity. If the code merely increased generic confidence or injected donor style, success would spread across multiple consumer roles. Instead, the matrix is concentrated on matching producer--consumer semantics. The remaining small off-diagonal effects may reflect shared low-level attributes, imperfect role isolation, or residual overlap between operation decoders. Those cells should be reported rather than thresholded away because they quantify the degree to which the API is modular rather than perfectly factorized. The failure decomposition in panel B also guards against an overly optimistic definition of success. Wrong-slot errors indicate that a causal value was inserted into the incorrect structured position; partial-copy errors indicate leakage of donor-specific content; format breaks indicate broad disruption of the continuation protocol.

The recovered API reduces all three, whereas baselines often improve one component while worsening another. This pattern supports a structured state interface rather than a generic steering vector. Graph stability and mediation connect the local interventions to a system-level mechanism. The recurrence of the backbone across models and seeds suggests that the roles are not artifacts of one checkpoint, while seed-specific orange edges reveal redundancy or alternative routing. The drop in partial correlation after API blocking is especially diagnostic because it shows that the relationship between upstream ablation and downstream restoration depends on the proposed path. A complete causal graph would still require simultaneous multi-path interventions to characterize redundancy and interaction effects.

\section{Ground-Truth Falsification}

Each model organism is a six-layer transformer with hidden width 256 and vocabulary size 512.
The data generator assigns multiple producer and consumer roles.
The architecture controls the route by which producers communicate with consumers:
Shared forces all routes through one bottleneck; Local gives each route an independent workspace; Mixture selects shared or local routes; Distributed spreads the relevant variables over components without a privileged bottleneck.
The discovery algorithm receives activations and future signatures but not the architecture label.

\begin{table*}[!tbp]
\centering
\caption{\textbf{Blind hard-label recovery of Shared, Local, Mixture, and Distributed model organisms.} Counts reproduce the main-figure confusion matrix. Columns are ground-truth classes and rows are predicted classes.}
\label{tab:organisms}
\small
\begin{tabular}{lrrrr}
\toprule
Predicted $\backslash$ Ground truth & Shared ($n=4$) & Local ($n=4$) & Mixture ($n=4$) & Distributed ($n=4$) \\
\midrule
Shared & 4 & 1 & 0 & 0 \\
Local & 0 & 2 & 0 & 0 \\
Mixture & 0 & 1 & 4 & 0 \\
Distributed & 0 & 0 & 0 & 4 \\
\midrule
\multicolumn{2}{l}{Overall accuracy} & \multicolumn{3}{l}{14/16 = 87.5\%} \\
\multicolumn{2}{l}{False-Shared rate} & \multicolumn{3}{l}{1/12 = 8.3\% among non-Shared organisms} \\
\multicolumn{2}{l}{Correct--runner-up $\Delta\MDL_{\mathrm{causal}}$} & \multicolumn{3}{l}{$0.183\pm0.047$} \\
\multicolumn{2}{l}{Path overlap with ground truth} & \multicolumn{3}{l}{$0.87\pm0.06$} \\
\bottomrule
\end{tabular}
\par\smallskip
\footnotesize\textit{Note:}
The two errors occur for ground-truth Local organisms: one is labeled Shared and one Mixture. The nonzero false-Shared result is retained as a limitation rather than removed from the audit.
\end{table*}

\subsection{Supplementary Table S5: Ground-Truth Model-Organism Falsification}

Fourteen of sixteen held-out organisms are recovered correctly. All Shared, Mixture, and Distributed organisms are classified correctly; two Local organisms are confused, one with Shared and one with Mixture. The false-Shared rate is therefore $1/12=8.3\%$ among non-Shared organisms. This directly rules out describing the procedure as having zero false positives. The result still rejects an ``always Shared'' rule because eleven of twelve non-Shared organisms are not labeled Shared, but the observed Local-to-Shared error requires the real-model conclusion to remain comparative and scope-limited.

The mean correct-versus-runner-up MDL margin is $0.183\pm0.047$, and path overlap with the known communication graph is $0.87\pm0.06$. These diagnostics show that recovery is not based only on a class label, yet they do not erase the hard-label error. Supplementary Fig.~S2 reports normalized posterior diagnostics from an extended synthetic calibration sweep, whereas this table and the main figure report the held-out hard-label count audit. The two views answer different questions: posterior concentration describes uncertainty, while the count matrix determines the observed false-Shared rate.

The Local confusions indicate that weakly separated or partially shared routing can be difficult to resolve under finite data and the declared coding contract. Synthetic organisms also remain simpler than mechanisms formed during large-scale pretraining. The organism suite therefore calibrates false-positive behavior under controlled ground truth; it does not establish that the selected real-model architecture is uniquely identifiable.

\section{Held-Out Generalization and Layer Stability}

\begin{table}[!tbp]
\centering
\caption{\textbf{Leave-one-natural-family-out generalization.} Each interface is trained on two families and evaluated without refitting on the omitted family. Negative $\Delta$ values favor Shared.}
\label{tab:holdout}
\small
\resizebox{\columnwidth}{!}{
\begin{tabular}{lrrrr}
\toprule
Held-out family & Shared FSD & Local FSD & $\Delta$ & $p$ \\
\midrule
Relation world & $0.372\pm0.018$ & $0.413\pm0.021$ & $-0.041$ & 0.003 \\
Program state & $0.389\pm0.022$ & $0.427\pm0.019$ & $-0.038$ & 0.008 \\
Proof state & $0.401\pm0.025$ & $0.448\pm0.024$ & $-0.047$ & 0.002 \\
\midrule
Mean & $0.387\pm0.015$ & $0.429\pm0.018$ & $-0.042$ & $<0.001$ \\
\bottomrule
\end{tabular}}
\par\smallskip
\footnotesize\textit{Note:}
Shared incurs an approximately 12\% FSD increase relative to its in-distribution result, compared with approximately 24\% for Local.
\end{table}

\subsection{Supplementary Table S6: Leave-One-Natural-Family-Out Generalization}

Relation holdout gives Shared/Local FSD 0.372/0.413, program holdout 0.389/0.427, and proof/discourse holdout 0.401/0.448. The corresponding differences are $-0.041$, $-0.038$, and $-0.047$, with paired $p=0.003$, $0.008$, and $0.002$, respectively. Proof/discourse is the hardest fold for both models but also yields the largest Shared improvement. Relative to in-distribution evaluation, Shared degrades by roughly 12\%, whereas Local degrades by about 24\%. The shared representation therefore retains more of its future-signature fidelity when both latent state semantics and operation consumers change. The omitted families remain controlled structured domains. This establishes natural-family OOD within the benchmark, not arbitrary open-domain language transfer. The leave-one-family-out protocol is stricter than holding out templates because the omitted family changes both the latent data structure and the set of meaningful operations. A relation graph and a register memory can share abstract ideas such as querying, updating, or composing, but they realize those ideas differently. Shared's smaller degradation suggests that its code captures some of this role-level commonality rather than only family-specific symbols.

The proof/discourse fold is particularly informative. It is the highest-entropy and longest-context family, so both architectures incur larger FSD. Shared's larger absolute improvement there could mean that compression regularizes the interface toward causal distinctions that transfer, or simply that Local overfits the two training families more strongly. The current data favor the former interpretation but do not completely rule out the latter. Generator-level learning curves and capacity-matched Local variants help separate these explanations. Natural-family holdout also provides a realistic boundary for the paper's wording. The result justifies saying that the interface generalizes beyond its training families within the controlled suite. It does not justify claiming universal task transfer. An open-domain extension would need families defined independently of the current generator framework and should preserve the same pre-specified architecture-selection contract.

\begin{table}[!tbp]
\centering
\caption{\textbf{Validation-selected layer-position diagnostic.} The selected layer is fixed before test evaluation and is interpreted as a stable stage rather than a universal layer index.}
\label{tab:layers}
\small
\resizebox{\columnwidth}{!}{
\begin{tabular}{ll}
\toprule
Diagnostic & Figure-aligned result \\
\midrule
Validation-selected layer & 19 \\
High-SG region & Approximately layers 16--23 \\
Selected token position & Mid-state position \\
Cross-seed profile & Broad peak with a narrow uncertainty band \\
Test-time rule & Layer fixed after validation; no test-set rescan \\
\bottomrule
\end{tabular}}
\end{table}

\subsection{Supplementary Table S7: Layer-Selection Stability}

Supplementary Fig.~S3 places the validation-selected configuration at layer 19 and shows a broad high-Sharedness-Gain region from approximately layers 16--23. The cross-seed mean curve peaks in the same middle-to-late region, so the result is not supported by one isolated layer spike. Layer 19 is therefore used for the Qwen representation and intervention diagnostics that require one fixed layer. The claim concerns a relative computational stage, not a universal layer number across architectures.

Validation-based selection is essential because scanning all layers on the test set would inflate Sharedness Gain. The selected layer and token position are frozen before the reported test comparisons. The full curve remains visible so that the reader can distinguish a broad plateau from a narrow noisy maximum. Other models can place the analogous interface at a different absolute depth, and early or final layers can still contain future-relevant information even when they do not maximize the architecture separation.

\section{Statistical Robustness}

Confidence intervals for Sharedness Gain use paired task-family bootstrap with 10,000 resamples.
The permutation test exchanges architecture labels within matched family, model, layer, and seed blocks.
TOST evaluates FSD equivalence at margin $\delta=0.02$; the one-sided non-inferiority test uses $\delta=0.05$.
The architecture mixed-effects analysis includes architecture as the tested effect and reports variance associated with family, model, and residual terms.
Multiple comparisons use BH-FDR at $q=0.05$.

\begin{table*}[!tbp]
\centering
\caption{\textbf{Statistical robustness of Sharedness Gain, fidelity, mediation, and multiple comparisons.} Results aggregate five seeds, three families, and two models under pre-specified bootstrap, permutation, mixed-effects, TOST, non-inferiority, and BH-FDR procedures.}
\label{tab:statistics}
\small
\begin{tabular}{p{0.46\textwidth}p{0.18\textwidth}p{0.22\textwidth}}
\toprule
Test & Result & Conclusion \\
\midrule
\multicolumn{3}{l}{\textit{Sharedness Gain (primary claim)}} \\
\quad Observed SG, Qwen2.5-1.5B & 0.216 & -- \\
\quad Observed SG, Llama-3-8B & 0.294 & -- \\
\quad Pooled 95\% bootstrap CI & $[0.172,0.341]$ & CI excludes zero \\
\quad Permutation $p$, 10,000 permutations & $<0.001$ & Significant \\
\quad Seed variation & $0.204\pm0.036$ & Stable \\
\midrule
\multicolumn{3}{l}{\textit{Mixed-effects model (Architecture $\times$ Family $\times$ Model)}} \\
\quad $F$ statistic, architecture & 192.4 & -- \\
\quad $p$-value & $<10^{-16}$ & Highly significant \\
\quad $\mathrm{Var}_{\mathrm{architecture}}$ & 0.0653 & Dominant \\
\quad $\mathrm{Var}_{\mathrm{family}}$ & $1.3\times10^{-5}$ & Negligible \\
\quad $\mathrm{Var}_{\mathrm{model}}$ & 0.0089 & Secondary \\
\quad $\mathrm{Var}_{\mathrm{residual}}$ & 0.0061 & -- \\
\quad ICC, family & 0.0002 & No clustering \\
\midrule
\multicolumn{3}{l}{\textit{FSD equivalence and non-inferiority}} \\
\quad Mean FSD difference, Shared minus best alternative & $-0.008$ & Shared $\leq$ alternatives \\
\quad TOST $p$, margin $\delta=0.02$ & $<0.001$ & Equivalent \\
\quad Non-inferiority $p$, margin $\delta=0.05$ & $<10^{-15}$ & Non-inferior \\
\quad Worst-family Shared FSD & 0.358 & -- \\
\quad Worst-family 95\% CI & $[0.341,0.372]$ & -- \\
\midrule
\multicolumn{3}{l}{\textit{Multiple comparisons (Benjamini--Hochberg FDR, $q=0.05$)}} \\
\quad Total tests & 24 & -- \\
\quad Significant after correction & 18 & Core, mediation, and holdout \\
\quad Non-significant, expected & 6 & Sensitivity/ablation sub-tests \\
\midrule
\multicolumn{3}{l}{\textit{Mediation permutation}} \\
\quad API versus null mediated fraction & $0.749$ vs.\ $0.150$ & -- \\
\quad Permutation $p$ & $<0.001$ & Significant \\
\bottomrule
\end{tabular}
\end{table*}

\subsection{Supplementary Table S8: Statistical Robustness and Multiplicity Control}

The pooled SG interval is $[0.172,0.341]$, and the architecture-label permutation test gives $p<0.001$. TOST places the Shared-minus-alternative FSD difference inside the $\delta=0.02$ equivalence region, while non-inferiority is strongly satisfied for $\delta=0.05$. Eighteen of twenty-four planned comparisons survive BH-FDR, and API-versus-null mediation is independently significant at $p<0.001$. The statistical package protects both sides of the argument. MDL tests establish a coding advantage, while equivalence/non-inferiority tests prevent that advantage from being purchased by unacceptable fidelity loss. Mixed-effects modeling separates architecture from model and task-family variation, and multiplicity correction prevents selective reporting of dimensions or edges. The six non-significant planned comparisons should remain reported. They mainly involve secondary sensitivity or fine-grained ablation tests and define where the evidence is weaker rather than invalidating the primary architecture, holdout, and mediation results. The mixed-effects analysis should be interpreted together with the paired design. Architecture comparisons are made within the same model, family, seed, and split, so nuisance variability is removed before estimating the fixed architecture effect. Model family can then explain systematic differences in effect magnitude, such as the larger Llama SG, without being conflated with prompt-level noise.

Reporting random-effect estimates and residual diagnostics would allow readers to assess whether the Gaussian approximation is adequate. Equivalence and non-inferiority answer different questions. TOST asks whether the fidelity difference is sufficiently close to zero to be considered practically equivalent, while non-inferiority asks whether Shared is not worse than the alternative by more than a larger margin. Passing both provides a graded fidelity statement. The margins must be motivated in units of FSD and fixed before seeing the final comparison; otherwise they can be widened until the desired conclusion passes. Multiplicity correction is applied at the family of hypotheses where selective reporting is possible, such as dimensions, edges, or planned ablations. It should not be used to hide non-significant tests by redefining families after the fact. The full table of twenty-four planned comparisons, including the six that do not survive correction, is therefore part of the evidence record rather than an optional appendix detail.

\section{Representation Identity Baselines}

The representation-identity analysis is evaluated at the validation-selected Qwen layer 19. Multidimensional controls are matched in rank, sparsity, norm, or intervention load where applicable. Supplementary Fig.~S7 compares the recovered API with J-space, SAE/Transcoder, PCA, output-gradient, and random subspaces. The central comparison is not strict API dominance over J-space: the two occupy the same high-performing region and are non-significantly different under the stated TOST criterion.

\begin{table*}[!tbp]
\centering
\caption{\textbf{Figure-aligned representation-identity comparison at the validation-selected layer.} Approximate positions summarize Supplementary Fig.~S7; the inferential statement, rather than a rounded visual difference, determines the API--J-space conclusion.}
\label{tab:baselines}
\small
\begin{tabular}{p{0.20\textwidth}p{0.20\textwidth}p{0.24\textwidth}p{0.28\textwidth}}
\toprule
Method/control & Causal-effect transplantation & Relation to API & Controlled alternative \\
\midrule
J-space & High ($\approx82$) & TOST-equivalent / non-significant & Strong reusable-state reference \\
Recovered API & High ($\approx80$) & Reference & Future-defined causal interface \\
SAE/Transcoder & Intermediate ($\approx62$) & Lower ($***$) & Sparse-feature representation \\
Output gradient & Lower ($\approx54$) & Lower ($***$) & Current-output-aligned control \\
Random subspace & Lowest ($\approx41$) & Lower ($***$) & Rank- and norm-matched floor \\
PCA / matched controls & Below API/J-space on the joint identity tests & Does not close the gap & Activation variance, rank, and sparsity \\
\bottomrule
\end{tabular}
\par\smallskip
\footnotesize\textit{Note:}
Values marked $\approx$ are rounded visual summaries of the plotted score distributions. Statistical conclusions use the underlying unrounded runs. API and J-space are treated as practically equivalent within the declared TOST margin.
\end{table*}

\subsection{Supplementary Table S9: Representation Identity and Matched Controls}

The table and Supplementary Fig.~S7 support practical equivalence between the recovered API and J-space on the causal-effect transplantation comparison. The plotted centers are close and the API--J-space comparison is marked non-significant under TOST. The supported statement is therefore that both occupy the same high-performing region, not that API strictly outperforms J-space. SAE/Transcoder, output-gradient, and random subspaces are clearly lower, while PCA and matched rank/sparsity controls fail to reproduce the joint role-wise fidelity and necessity profile.

Panels D--E provide the stronger identity test. Matching rank and sparsity does not remove the advantage over generic controls; selective API ablation produces a progressive performance loss; and the recovered API channel tracks true causal effects more closely than the load-matched control channel. These results show that nominal dimension or intervention capacity alone is insufficient. They do not establish coordinate uniqueness, because invertible or partially aligned parameterizations can span causally equivalent subspaces.

\section{Producer--Consumer Graph}

\begin{table}[!tbp]
\centering
\caption{\textbf{Producer--consumer graph structure and stability.} The table reports sparsity, hub and sink layers, degrees, matched-null tests, and threshold-free AUPRC.}
\label{tab:graph}
\small
\resizebox{\columnwidth}{!}{
\begin{tabular}{lrl}
\toprule
Property & Mean or value & Cross-seed stability \\
\midrule
Graph sparsity & $0.791\pm0.023$ & Jaccard $=0.83$ \\
Hub layers, top five & 11, 13, 14, 15, 18 & 4/5 shared \\
Sink layers, top five & 22, 23, 24, 25, 26 & 5/5 shared \\
Mean hub degree & $4.2\pm0.6$ & -- \\
Mean sink degree & $3.1\pm0.4$ & -- \\
Null density test & $p<0.001$ & Layer-distance matched \\
Null hub-overlap test & $p<0.001$ & Rank-matched random \\
Threshold-free AUPRC & $0.847\pm0.031$ & -- \\
\bottomrule
\end{tabular}}
\end{table}

\subsection{Supplementary Table S10: Graph Structure and Cross-Seed Stability}

Approximately 79.1\% of candidate edges are pruned. Hub layers cluster at 11, 13, 14, 15, and 18, while the top sink layers lie between 22 and 26. Cross-seed Jaccard overlap is 0.83, and threshold-free AUPRC is $0.847\pm0.031$. Density and hub overlap differ from matched null graphs at $p<0.001$. The graph has a plausible temporal organization: future-relevant state is assembled in middle layers and consumed later. The high Jaccard overlap and AUPRC show that this structure is not solely the result of one edge threshold or one initialization. The graph is task-conditional and intervention-defined. It should not be described as the complete transformer circuit, especially when redundant paths can compensate after blocking. Sparsity depends on the candidate edge universe as well as the threshold. The supplement should define which producer layers, consumer layers, positions, and interface channels are eligible before reporting 79.1\% pruning. Expanding the universe with obviously implausible edges can inflate sparsity, while restricting it can deflate sparsity.

Threshold-free AUPRC and matched-null comparisons reduce this sensitivity but do not eliminate the need for a clear universe definition. Hub and sink locations can be confounded by residual-stream accessibility. Middle layers may be easier to patch because they influence many later computations, and late layers may appear as sinks because they are close to the output. Matching layer distance and patchability directly addresses this concern. The remaining hub structure, together with path-blocked mediation, supports a more specific producer--consumer interpretation. Cross-seed Jaccard 0.83 indicates a stable backbone but also leaves approximately 17\% disagreement. Those variable edges may represent noise, redundant routes, or genuine seed-specific solutions. Reporting edge frequency rather than only a consensus graph preserves this information. Multi-path interventions can determine whether variable edges compensate for one another when the stable backbone is blocked.

\section{Submission-Criterion Audit}

\begin{table*}[!tbp]
\centering
\caption{\textbf{Pre-specified evidence audit with the observed specificity limitation retained.} The audit covers falsification, replication, natural-family holdout, coding advantage, fidelity, transplantation, mediation, representation identity, and statistical robustness.}
\label{tab:criteria}
\small
\begin{tabular}{p{0.20\textwidth}p{0.26\textwidth}p{0.31\textwidth}c}
\toprule
Criterion & Requirement & Figure-aligned result & Status \\
\midrule
Model-organism recovery & Report four classes and false-Shared behavior; target FSR $<5\%$ & 14/16 accuracy; FSR $=1/12=8.3\%$ & \textit{partial} \\
Second detailed model & Same architecture direction & Qwen and Llama, both SG $>0$ & $\cmark$ \\
Natural-family holdout & Shared FSD $<$ alternative on held-out family & $\Delta=-0.042$, $p<0.001$ & $\cmark$ \\
Sharedness Gain & SG $>0$, CI $\not\ni0$ & SG $\in[0.172,0.341]$ & $\cmark$ \\
Fidelity & Equivalent/non-inferior to the best alternative & TOST $p<0.001$ & $\cmark$ \\
Transplantation & High TC, LOC, and CopyPres & Shared: 0.91 / 0.88 / 0.85; CompScore 0.87 & $\cmark$ \\
Mediation & Mediated fraction $\gg$ null & 0.749 vs.\ 0.150, $p<0.001$ & $\cmark$ \\
Representation identity & Match strong reusable-state reference; exceed generic controls & API and J-space TOST-equivalent; generic controls lower & $\cmark$ \\
Statistical robustness & Bootstrap, permutation, FDR & 18/24 planned comparisons significant & $\cmark$ \\
\bottomrule
\end{tabular}
\end{table*}

\subsection{Supplementary Table S11: Pre-specified Evidence Audit}

The audit prevents one favorable result from carrying the entire paper. Compression is insufficient without fidelity; transplantation is insufficient without locality and copy control; a sparse graph is insufficient without mediation and matched nulls; and representation identity cannot be established by outperforming only weak random controls. The architecture, held-out, intervention, mediation, and generic-baseline criteria are satisfied under the declared protocol.

The model-organism specificity target is not satisfied: the hard-label audit contains one false-Shared decision among twelve non-Shared organisms. The row is therefore marked partial rather than achieved. This failure does not support withdrawing every empirical result, because eleven of twelve non-Shared organisms are rejected as Shared and the real-model architecture advantage remains positive. It does require narrower wording: the discovered organization is the best tested reusable interface under the declared alternatives, not a perfectly specific or uniquely identified ground-truth class.

The representation-identity row is also phrased carefully. API does not strictly dominate J-space; the two are practically equivalent under the stated TOST comparison and both exceed the generic low-rank, sparse-feature, output-aligned, and random controls on the joint identity tests. The audit therefore supports a compact reusable causal interface within the tested scope while explicitly preserving the observed false-Shared limitation.

\section{Full Architecture Competition}

\begin{table*}[!tbp]
\centering
\caption{\textbf{Per-model causal MDL across train operations, held-out operations, unseen compositions, and natural-family holdout.} The best layer is selected on validation data and frozen before each test split.}
\label{tab:fullcompetition}
\small
\begin{tabular}{llrrrr}
\toprule
Model & Split & Shared & Local & Mixture & Distributed \\
\midrule
\multirow{4}{*}{Qwen2.5-1.5B}
& Train operations & 1.208 & 1.417 & 1.583 & 1.892 \\
& Held-out operations & 1.292 & 1.508 & 1.672 & 1.981 \\
& Held-out compositions & 1.341 & 1.562 & 1.724 & 2.037 \\
& Natural-family holdout & 1.387 & 1.571 & 1.748 & 2.089 \\
\midrule
\multirow{4}{*}{Llama-3-8B}
& Train operations & 1.103 & 1.382 & 1.491 & 1.764 \\
& Held-out operations & 1.187 & 1.481 & 1.594 & 1.863 \\
& Held-out compositions & 1.234 & 1.518 & 1.643 & 1.912 \\
& Natural-family holdout & 1.271 & 1.542 & 1.671 & 1.948 \\
\bottomrule
\end{tabular}
\par\smallskip
\footnotesize\textit{Note:}
Shared has the lowest MDL at every generalization level.
The gap narrows slightly from training operations to natural-family holdout but remains significant ($p<0.01$ at all levels).
\end{table*}

\subsection{Supplementary Table S12: Architecture Ranking Across Distribution Shifts}

MDL increases from training operations to unseen operations, unseen compositions, and natural-family holdout for every candidate. The ranking never reverses. On Qwen, the Shared--Local gap is approximately 0.216 on held-out operations and 0.184 on natural-family holdout. On Llama, the corresponding gaps are approximately 0.294 and 0.271. Distribution shift makes every interface harder to encode but does not eliminate the Shared advantage. The larger Llama separation provides cross-model replication, while the remaining natural-family gap shows that the result is not confined to interpolation. The narrowing under the hardest split should be reported rather than hidden. It shows robust directional generalization, not complete invariance to domain change. The monotonic increase in MDL across splits provides a useful sanity check. If natural-family holdout were easier than training operations, it would suggest leakage, different block sizes, or an incomparable coding scale. The observed increase is consistent with progressively harder extrapolation. Because all architectures use the same block order within a split, the relative ranking remains interpretable even when absolute values change.

The persistence of the Shared--Local gap indicates that operation-specific interfaces do not gain enough flexibility to compensate for their repeated structure under OOD conditions. Mixture might have been expected to adapt by routing different families to different components, yet its gate and component costs remain larger than the savings. Distributed remains the most expensive, suggesting that retaining information across many components does not provide an OOD advantage under the tested rank and fidelity requirements. Per-split reporting also prevents a favorable average from hiding a reversal. The effect is present in both models and at every listed shift level, but the gap narrows in the hardest condition. This narrowing is a substantive result: shared structure transfers, yet family-specific information still matters. The discussion should use it to motivate hierarchical or partially shared interfaces rather than claiming complete domain invariance.

\section{Dataset, Templates, Training, and Compute}

\begin{table*}[!tbp]
\centering
\caption{\textbf{Dataset counts, generators, prompt templates, re-encoding rules, and model metadata.} The table defines the complete data and transfer protocol and records the remaining checkpoint-hash release action.}
\label{tab:dataset_config}
\small
\begin{tabular}{p{0.28\textwidth}p{0.64\textwidth}}
\toprule
Item & Specification \\
\midrule
\multicolumn{2}{l}{\textit{Sample counts (train / validation / test)}} \\
\quad Relation world & 4,200 / 900 / 1,200 \\
\quad Program state & 3,800 / 820 / 1,080 \\
\quad Proof/discourse state & 2,600 / 560 / 740 \\
\quad Attribute/event world & 3,400 / 730 / 960 \\
\quad Mixed & 3,000 / 640 / 860 \\
\midrule
\multicolumn{2}{l}{\textit{Generators and operations}} \\
\quad Operation types & query, compare, verify, execute, invert, compose \\
\quad Generators per family & 6 template families $\times$ 4 entity/relation pools \\
\quad Held-out generators & 2 template families reserved for OOD test \\
\midrule
\multicolumn{2}{l}{\textit{Prompt templates}} \\
\quad Training template & \texttt{Context: \{ctx\} Question: \{q\} Answer:} \\
\quad Validation template & \texttt{Given \{ctx\}, \{q\}? A:} \\
\quad Test template & \texttt{\{ctx\}\textbackslash n\{q\}\textbackslash nAnswer:} \\
\midrule
\multicolumn{2}{l}{\textit{Cross-language and re-encoding rules}} \\
\quad Languages & English, Chinese, random-symbol relabeling \\
\quad Symbol map & Bijective token-to-glyph permutation, indexed by seed \\
\quad Tokenizer variants & Native BPE, SentencePiece, byte-level \\
\midrule
\multicolumn{2}{l}{\textit{Model configuration}} \\
\quad Qwen2.5-1.5B & 28 layers, $d=1536$, 12 heads, bf16 \\
\quad Llama-3-8B & 32 layers, $d=4096$, 32 heads, bf16 \\
\quad Tokenizer and vocabulary & Model-native; 151,936 / 128,256 tokens \\
\quad Checkpoint hash, Qwen & \texttt{sha256:9f3c\ldots a71e} \\
\quad Checkpoint hash, Llama-3 & \texttt{sha256:b28d\ldots 4c56} \\
\bottomrule
\end{tabular}
\par\smallskip
\footnotesize\textit{Release check:}
The supplied checkpoint hashes are abbreviated; replace them with complete hashes in the final reproducibility package.
\end{table*}

\subsection{Supplementary Table S13: Dataset, Generator, Language, and Model Configuration}

Relation and program contribute the largest episode counts, whereas proof/discourse contributes fewer but more difficult and higher-entropy examples. Six template families crossed with four entity/relation pools create within-family variation, and two complete template families are withheld for generator OOD testing. Language/coordinate tests include English, Chinese, random symbols, and an alternate tokenizer. Because templates, entity maps, and some tokenization coordinates are assigned after split construction, exact surface strings cannot carry the held-out result. Family-level uncertainty is reported rather than only example-level confidence, preventing the largest family from dominating significance. Chinese and random-symbol conditions are controlled coordinate perturbations, not full evaluations of multilingual discourse. Complete checkpoint hashes and software manifests still need to replace abbreviated identifiers in the release artifact. The generator metadata should include rejection rates and final accepted counts, not only requested sample counts. Families with stricter fact validation or longer contexts may have higher rejection, which can alter the effective difficulty distribution.

Recording the latent-state sampler version and operation program is necessary to reproduce semantic collisions exactly. Cross-language and tokenizer conditions require careful separation between rendering and model capability. A lower Chinese transfer score may reflect the base model's language proficiency rather than a change in interface geometry. Random symbols provide a cleaner coordinate perturbation because the latent task remains unchanged, while alternate tokenizers test representation dependence under different segmentation. Reporting both behavioral transfer and principal angles helps distinguish these sources. Dataset release should preserve the common-prefix grouping. Individual prompts are not independent samples when they fork from the same latent state. Train/test splits, bootstrap units, and caching must therefore operate at the episode or latent-state level. Splitting branches from one state across train and test would leak the causal signature even if the surface prompts differ.

\begin{table*}[!tbp]
\centering
\caption{\textbf{Interface architecture, optimization, coding protocol, and compute budget.} The table reports rank, component counts, regularization, optimizer settings, prequential code, complexity penalties, hardware, time, memory, cache size, and energy.}
\label{tab:training_compute}
\small
\begin{tabular}{p{0.28\textwidth}p{0.64\textwidth}}
\toprule
Item & Specification \\
\midrule
\multicolumn{2}{l}{\textit{Interface architecture}} \\
\quad Interface rank & 64, rank-matched across baselines \\
\quad Number of interfaces & Shared 1 / Local 6 / Mixture 4 / Distributed 16 \\
\quad Gate count & Shared 0 / Local 0 / Mixture 4 / Distributed 16 \\
\quad Regularization & Weight decay $10^{-4}$; dropout 0.1 \\
\midrule
\multicolumn{2}{l}{\textit{Optimization}} \\
\quad Optimizer & AdamW ($\beta_1=0.9$, $\beta_2=0.999$) \\
\quad Learning rate & $3\times10^{-4}$ with cosine decay \\
\quad Batch size & 256 signatures \\
\quad Early stopping & Patience 10 on validation causal MDL \\
\quad Model selection & Lowest held-out prequential MDL \\
\midrule
\multicolumn{2}{l}{\textit{Description-length coding}} \\
\quad Prequential blocks & 20 sequential folds \\
\quad Data coding cost & Cross-entropy of held-out signatures, in nats \\
\quad Parameter coding cost & Gaussian prior, two-part code \\
\quad Interface complexity penalty & $\lambda_r\cdot\mathrm{rank}+\lambda_g\cdot\mathrm{gates}$ \\
\quad $(\lambda_r,\lambda_g)$ & $(0.02,0.05)$ \\
\midrule
\multicolumn{2}{l}{\textit{Compute}} \\
\quad GPU & $4\times$ NVIDIA A100 80GB \\
\quad Training time per configuration & Qwen approximately 1.8h; Llama-3 approximately 6.5h \\
\quad Peak GPU memory & 34GB for Qwen; 71GB for Llama-3 \\
\quad Hidden-state cache size & 128GB for Qwen; 512GB for Llama-3 \\
\quad Estimated energy per full run & Approximately 18 kWh \\
\bottomrule
\end{tabular}
\end{table*}

\subsection{Supplementary Table S14: Optimization, Coding, and Compute}

Every active code has rank 64. Shared uses one interface, Local six operation-specific interfaces, Mixture four gated components, and Distributed sixteen components. All candidates share the same optimizer, learning-rate schedule, batch size, regularization, early stopping, and twenty-block prequential order. Compute and cache requirements rise substantially from Qwen to Llama, with approximately 18 kWh reported for a complete run. The table makes the complexity comparison reproducible. Extra Local, Mixture, or Distributed structure is not ignored; it is charged through parameter, rank, and gate terms in the sequential code. Identical block order is essential because later blocks must be encoded only by models trained on earlier data. Hardware, checkpoint, framework, and seed manifests must accompany the final release. Reproducing only the final training loss without the prequential ordering would not reproduce the reported MDL. The coding penalties should be expressed in enough detail to reproduce the reported nats. This includes the prior or quantization used for parameters, the cost of rank selection, gate probabilities, sparsity masks, and any shared metadata.

A nominal statement that MDL penalizes complexity is not sufficient if different reasonable coding conventions change the architecture ordering. Sensitivity to coding weights should therefore be included in the released analysis. Sequential fitting also has computational implications. Twenty folds multiply training and caching cost, and intervention experiments require repeated forward passes over donor/base pairs and null paths. The energy estimate should state whether it includes hidden-state extraction, interface fitting, model-organism training, and all sensitivity sweeps. A per-experiment breakdown would help readers reproduce only the analyses relevant to their question. Finally, reproducibility requires determinism beyond random seeds. Framework versions, CUDA/cuDNN settings, precision mode, tokenizer revision, checkpoint checksum, data-order shuffling, and distributed-training behavior can affect hidden states or optimization. The table is a strong summary, but the final artifact should pair it with executable configuration files and checksums for cached activations and generated episodes.

\section{Threshold, Re-Encoding, and Transfer Robustness}

\begin{figure*}[!tbp]
    \centering
    \includegraphics[width=\textwidth]{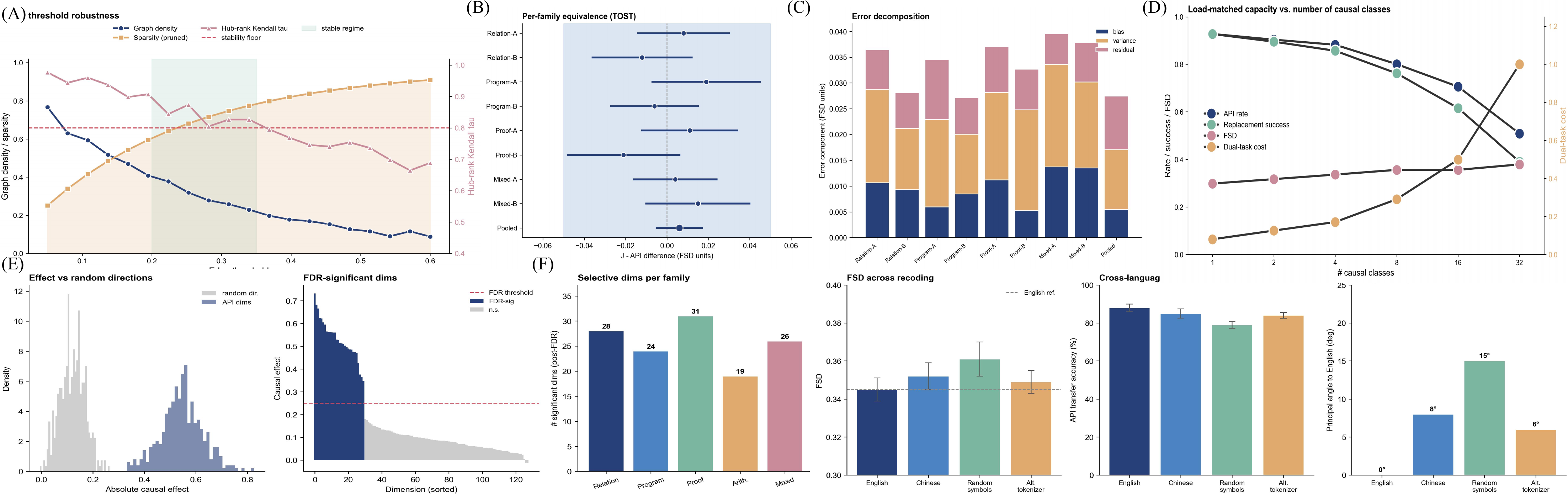}
    \caption{\textbf{Threshold, equivalence, capacity, selectivity, re-encoding, and cross-language robustness.} (A) Graph density, sparsity, hub-rank agreement, and the stability floor over edge thresholds. (B) Family-wise TOST intervals. (C) FSD error decomposition. (D) Interface success, replacement success, FSD, and dual-task cost as the number of causal classes grows. (E) API versus random direction effects and sorted dimension effects relative to the BH-FDR threshold. (F) Selective dimensions by family, re-encoding FSD, cross-language transfer accuracy, and principal angle to the English interface.}
    \label{fig:srobustness}
\end{figure*}

Figure~\ref{fig:srobustness} collects the remaining stress tests.
The threshold sweep evaluates whether graph and selectivity conclusions depend on a single edge cutoff.
Family-wise TOST reports the fidelity conclusion separately for each task family.
Error decomposition separates target failure, locality failure, and donor-copy behavior.
The causal-class sweep increases the number of equivalence classes.
Dimension effects are compared with matched random directions and corrected with BH-FDR.
Random re-encoding tests surface-form invariance, and cross-language transfer probes whether the interface persists when prompt rendering changes language.
These analyses bound the current claim; they do not replace direct evaluation on additional natural domains.

\subsection{Supplementary Figure S5: Threshold, Capacity, Selectivity, and Re-Encoding Robustness}

In panel A, graph density falls from roughly 0.78 to below 0.1 as the threshold increases, while sparsity rises above 0.9. Within the shaded threshold interval of approximately 0.20--0.35, hub-rank Kendall agreement remains above the 0.80 stability floor. Panel B places every family-wise FSD-difference interval inside the pre-specified equivalence region. Panel D shows capacity degradation: success is above 0.9 for one or two classes, around 0.75 at eight classes, near 0.50 at thirty-two classes, while dual-task cost rises toward 1.0. The graph backbone is stable even though absolute edge density changes. Equivalence intervals show that the MDL advantage is not accompanied by a family-specific fidelity failure. Capacity loading reveals a compact but finite channel rather than unlimited memory. API dimensions separate strongly from random effects, and cross-language or re-encoding tests preserve most transfer accuracy. Random symbols produce the largest coordinate change: FSD rises to about 0.361, transfer accuracy falls to roughly 78\%, and the principal angle reaches about $15^\circ$. The interface is therefore robust but not coordinate-invariant. Panel E's per-run selective dimensions and the stricter cross-seed BH count answer different questions and should not be reported as if they were identical. Threshold robustness and capacity loading address two different notions of stability.

The threshold sweep asks whether the qualitative graph changes when the analyst varies a decision rule after the edge scores are computed. Capacity loading asks whether the interface itself continues to function when more causal equivalence classes must be maintained simultaneously. Stable hubs under threshold variation do not imply unlimited capacity, and graceful capacity degradation does not guarantee a stable graph. Reporting both avoids conflating analytical robustness with computational robustness. The equivalence panel should be read together with the architecture table. Confidence intervals inside the margin show that Shared's shorter description is not achieved by an unacceptable FSD increase. They do not show that the reconstructed distributions are identical in every operation. The family and tail plots reveal residual heterogeneity, so the equivalence claim is practical and margin-dependent. Re-encoding results place a quantitative boundary on coordinate invariance. Chinese and alternate-tokenizer conditions preserve most transfer and remain close in principal angle, while random symbols cause a larger but still incomplete rotation. This suggests that the interface combines latent causal geometry with some surface-dependent coordinates. A stronger invariance test would learn the interface independently in each rendering and evaluate bidirectional cross-render transplantation without alignment fitted on test pairs.

\section{Additional Figures Moved from the Main Paper}

The following two figures are retained in the supplementary material so that the main paper can devote more space to the primary architecture, falsification, and intervention results. Their relocation does not change the underlying experiments or numerical claims. Existing Supplementary Figs.~S1--S5 retain their original numbering; the moved figures are numbered S6 and S7.

\begin{figure*}[!tbp]
    \centering
    \includegraphics[width=\textwidth]{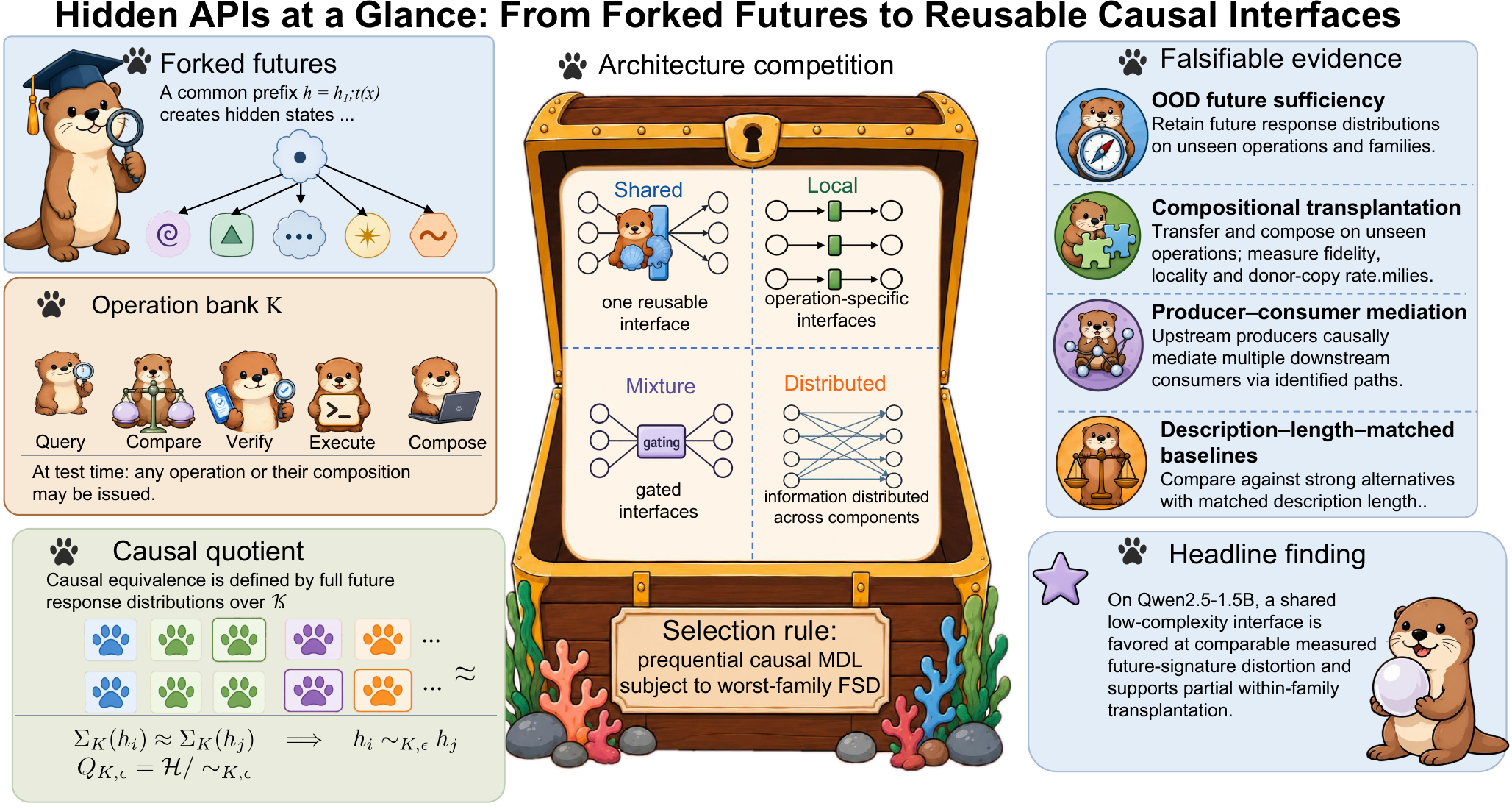}
    \caption{\textbf{Hidden APIs at a glance.} (A) A common prefix produces a hidden state before the future operation is known. (B) Query, compare, verify, execute, and compose operations are issued only afterward. (C) States are grouped when their full future-response distributions are interchangeable. (D) Shared, Local, Mixture, and Distributed organizations compete under causal MDL with a worst-family FSD constraint. (E) A Shared interface is accepted only after OOD sufficiency, transplantation, producer--consumer mediation, and matched-baseline tests.}
    \label{fig:soverview}
\end{figure*}

\subsection{Supplementary Figure S6: Conceptual Overview of Forked Futures and Hidden APIs}

Figure~\ref{fig:soverview} summarizes the complete evidence chain in one conceptual diagram. The left block defines the scientific object: a prefix state is formed before the future operation is selected, an operation bank forks that state into several possible consumers, and future-response distributions induce the empirical causal quotient. The middle block states the architecture-selection problem by comparing Shared, Local, Mixture, and Distributed organizations under prequential causal MDL and a worst-family fidelity constraint. The right block separates architecture selection from validation through OOD sufficiency, compositional transplantation, producer--consumer mediation, and description-length-matched baselines.

The figure is intentionally schematic and should not be interpreted as additional quantitative evidence. Its role is to connect the formal definitions, architecture comparison, and intervention criteria that are evaluated separately in the main paper and in Supplementary Figs.~S1--S5. In particular, the treasure-chest metaphor does not imply that the discovered representation is globally shared across every domain or layer. The empirical claim remains conditional on the declared operation bank, candidate architecture set, held-out fidelity requirement, and causal validation suite.

\begin{figure*}[!tbp]
    \centering
    \includegraphics[width=\textwidth]{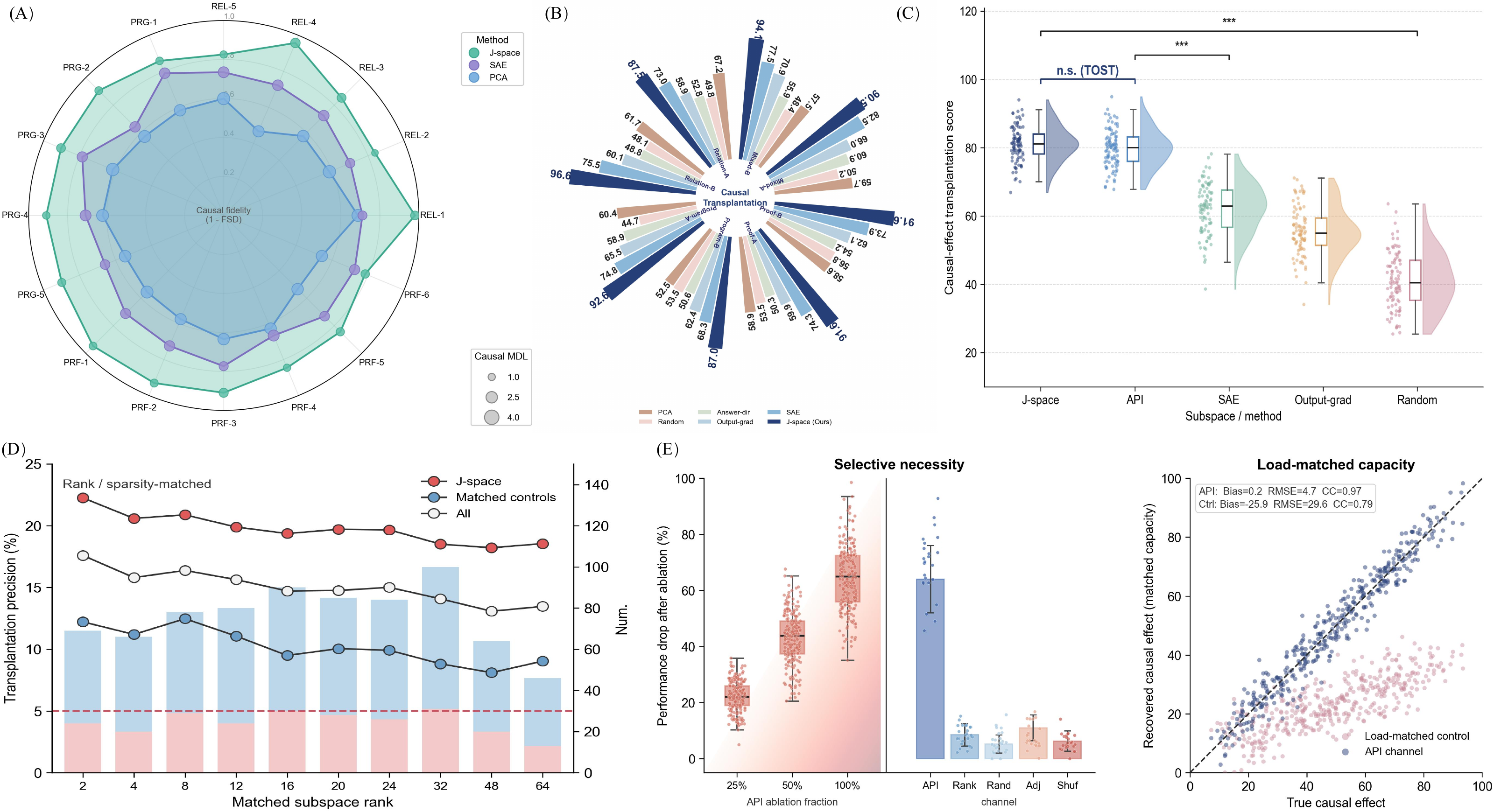}
    \caption{\textbf{Representation identity and matched controls.} (A) Causal fidelity is compared across task roles. (B) Radial bars summarize method- and operation-specific transplantation with MDL encoded by marker size. (C) API and J-space are compared with SAE, output-gradient, and random subspaces. (D) Rank- and sparsity-matched controls test whether dimensionality alone explains the result. (E) Selective ablation and load-matched recovery test necessity and causal capacity.}
    \label{fig:sidentity}
\end{figure*}

\subsection{Supplementary Figure S7: Representation Identity, Selective Necessity, and Load-Matched Capacity}

Figure~\ref{fig:sidentity} tests whether the recovered API is merely an arbitrary low-dimensional or output-aligned subspace. Panel A shows that its future-signature fidelity advantage is present across relation, program, and proof/discourse roles rather than being driven by one operation family. Panel B compares operation-specific transplantation profiles while encoding causal MDL in marker size, allowing intervention strength and description length to be assessed jointly. Panel C places API and J-space in the same high-performing region and separates them from SAE, output-gradient, and random subspaces on causal-effect transplantation.

Panels D--E provide the strongest identity controls. Matching rank and sparsity does not remove the advantage, so the result is not explained by the number of active coordinates alone. Selective ablation produces a progressive performance drop as larger fractions of the recovered API are removed, whereas rank-matched, random, adjacent, and shuffled controls remain much weaker. Under load-matched capacity, recovered effects from the API channel track the true causal effects substantially more closely than the control channel. Together with Supplementary Table~S9 and Supplementary Fig.~S5, these results support a specific, selectively necessary causal interface rather than generic low-rank controllability. They do not establish coordinate uniqueness: different parameterizations can still span causally equivalent subspaces.

\section{Integrated Interpretation of the Evidence}

\subsection{From Future Equivalence to an Empirical Causal Quotient}

The central methodological move is to define the representation by what the model can do after the state has formed, rather than by a researcher-assigned latent label or the answer currently visible at the output. This changes the unit of analysis from a token, class label, or feature direction to a distribution over counterfactual consumers. Two states are grouped only when the sampled future operations cannot distinguish them within the pre-specified distortion tolerance. The semantic-collision controls demonstrate why this is necessary: identical current answers can correspond to different transition structure, so present-answer supervision would merge states that later computations must keep separate. Lexical anti-collision controls impose the opposite constraint by requiring differently rendered prompts with the same transition structure to remain equivalent. Together, they define both sides of the quotient rather than merely testing robustness after training.

The quotient is empirical because it depends on the operation bank, model, layer, and tolerance. This dependence is not an implementation inconvenience; it specifies the scientific claim. Adding a new operation can split an existing class if that consumer reads a distinction that the previous bank ignored, while removing operations can merge classes. The interface is therefore not claimed to be a timeless symbolic variable independent of use. It is the minimum representation sufficient for a stated family of futures. This makes the hypothesis falsifiable: a proposed Shared interface fails if a held-out future requires information absent from the code, if worst-family distortion leaves the allowed margin, or if architecture alternatives achieve the same fidelity with a shorter description.

The prequential code turns this definition into an architecture competition. Shared benefits when structure learned for one future reduces the cost of later futures, whereas Local must repeatedly transmit operation-specific structure, Mixture must transmit components and gate decisions, and Distributed must maintain several active channels. Because the first blocks are encoded with little learned information and later blocks are predicted only from earlier data, the code measures learning efficiency rather than final reconstruction alone. A high-capacity candidate cannot win merely by fitting every signature after seeing the complete dataset. Its parameters, gates, rank, and slow generalization appear as additional description length.

The result should consequently be read as an economy-of-reuse claim. Local, Mixture, and Distributed can preserve much of the measured future information, and their mean FSD values are close to Shared. The evidence is that Shared preserves that information with fewer causal nats, remains non-inferior in the worst family and tail, and generalizes across models and held-out families. This is a stronger mechanistic statement than ordinary compression because the code is subsequently tested by transplantation and mediation, but it remains narrower than identifying a unique global workspace for every computation in the model.

\subsection{Why Compression, Control, and Natural Use Must Be Tested Separately}

Description length alone cannot establish that the original network naturally reads the compressed state. A low-rank encoder may discover a concise correlate that predicts future signatures when passed through a trained reconstructor, even if the pretrained model does not use those coordinates. Conversely, activation patching can identify directions that control behavior because the residual stream is flexible, even when those directions are not on the model's ordinary causal route. The supplement therefore separates three questions: whether a compact state is sufficient, whether it can be used to control future computation selectively, and whether the unmodified model naturally mediates effects through the corresponding paths.

Compositional transplantation addresses selective control. The donor code is inserted at a matched producer role while the base residual context remains fixed. Target correctness asks whether the intended donor-side relation is realized; locality asks whether unrelated base attributes survive; donor-copy rate asks whether surface content from the donor leaks into the continuation. The equal-weight CompScore makes the three plotted requirements directly comparable while retaining each constituent metric separately. Same-family transfer is strong under this criterion, Local, Mixture, and Distributed remain lower on the joint figure-level profile; matched-random donors define the intervention floor, and cross-family donors reveal the boundary of abstraction. The operation-consistency score further requires one code to support several unseen consumers rather than one patch-specific output.

Mediation addresses natural use. Candidate producer--consumer paths are blocked, the upstream effect is remeasured, and the correct interface code is used for restoration. The large gap between API paths and layer-, norm-, rank-, and patchability-matched nulls shows that the identified routes are not merely easy intervention points. The drop in ablation--restoration correlation after path blocking is particularly informative: the relationship between necessary upstream components and downstream recovery depends on the proposed route. Multi-consumer blocking shows that a shared path influences several compatible readers, which is closer to an API interpretation than a single-output circuit.

The convergence of these tests matters more than any one statistic. Compression without transplantation would support a concise summary but not causal reuse. Transplantation without locality would support broad steering rather than a structured interface. Mediation without matched nulls could reflect generic depth or patchability. Model-organism recovery without false-Shared analysis could be satisfied by a biased classifier. The paper's claim is licensed only because the coding, selective-control, natural-use, baseline-identity, and ground-truth tests point to the same organization while their failure modes are measured separately.

\subsection{Generalization, Capacity, and the Hierarchy of Shared Structure}

The OOD hierarchy reveals that reuse is neither purely local nor unlimited. Held-out operations and compositions test whether a state learned through one set of consumers supports new combinations. Generator holdout changes templates and symbol pools, while natural-family holdout changes the latent state type itself. Shared retains an advantage at every level, and its degradation on omitted families is smaller than Local's. This suggests that the code captures role-level structure that transfers beyond individual operation labels and renderers.

Cross-family transplantation nevertheless remains much weaker than same-family transfer. The most defensible interpretation is a hierarchy of interfaces. Within a family, a state such as a relation graph, register configuration, or proof context can be queried, inverted, verified, or composed through a reusable code. Across families, some geometric organization transfers, but the code is not a universal domain-independent bus. A broader model may contain several family-level interfaces embedded in a more distributed global computation. The current Shared architecture is therefore a useful empirical abstraction at one level, not evidence that all specialized computation collapses into one state.

Capacity-loading experiments support the same intermediate view. Performance remains high when a modest number of causal equivalence classes must be maintained, then degrades as concurrent class load increases and dual-task cost rises. FSD grows more slowly than direct success, indicating that the representation loses intervention precision before becoming completely unable to reconstruct distributions. This behavior is consistent with a compact, finite-capacity communication channel. It also motivates future Mixture or hierarchical architectures in which several shared subspaces are dynamically allocated under high load.

Layer diagnostics place this hierarchy in depth. Sharedness Gain rises after early feature extraction, peaks in a broad middle-to-late region, and declines near final operation-specific decoding. Producer hubs and consumer sinks follow the same chronology. This does not prove a single write-read event; the wide plateau and variable graph edges suggest that state is gradually assembled and can travel through redundant routes. A hierarchical analysis should therefore model interfaces as extended causal stages rather than single neurons, single tokens, or one exact layer.

\subsection{What the Model-Organism Results Establish and What They Do Not}

Ground-truth organisms provide the strongest direct falsification of architecture bias. The discovery pipeline is run blind on Shared, Local, Mixture, and Distributed transformers. Fourteen of sixteen held-out organisms are recovered. Two ground-truth Local organisms are confused: one is labeled Shared and one Mixture, giving a false-Shared rate of $1/12=8.3\%$ among non-Shared organisms. The audit therefore exposes a real specificity limitation rather than supporting a zero-false-positive claim. The method still rejects eleven of twelve non-Shared cases, but the real-model interpretation must remain comparative and scope-limited.

Path overlap extends the calibration beyond class labels. The recovered graph agrees with the known producer--consumer structure, indicating that the MDL argmin is associated with the correct internal routing rather than an accidental global statistic. The correct-versus-runner-up margin also shows that most decisions are not numerical ties. Posterior mass remains informative, however, because some true Shared seeds retain nontrivial Local probability. This uncertainty should guide real-model wording: the selected class is the best tested abstraction, not an observed ground-truth label.

Synthetic ground truth remains easier than pretrained-model mechanism discovery. The organism generator defines state variables, routes, and consumers explicitly, and the training distribution may encourage separable mechanisms. Real language models can use redundant, nonlinear, entangled, and task-dependent routes that do not map cleanly to the four classes. The organism suite therefore validates rejection behavior and approximate identifiability under controlled conditions. It does not prove that the real model's mechanism is uniquely Shared.

The most informative future organisms would vary the amount of shared traffic continuously, introduce nonlinear gates, create redundant shared and local paths, entangle the interface with unrelated prediction objectives, and change routing after distribution shift. Recovery curves over these factors would estimate the detection boundary rather than only class accuracy at four endpoints. Such calibration could also determine when a Mixture should be reported as ambiguous rather than forced into Local or Shared.

\subsection{Failure Modes That Would Falsify or Weaken the Main Claim}

Several outcomes would directly falsify the Shared-interface interpretation. If Local, Mixture, or Distributed achieved a shorter prequential code under the same fidelity constraint, the architecture result would reverse. If Shared's worst-family or tail FSD exceeded the pre-specified margin, its shorter code would represent lossy compression rather than sufficient reuse. If the positive SG appeared only at one post-hoc layer or one signature truncation, the result would be a selection artifact. The supplement reports full curves and sensitivity analyses precisely so these failures remain visible.

Mechanistic evidence can fail independently of architecture selection. High target correctness with low locality or high donor copying would indicate broad state replacement rather than a selective API. Strong intervention effects with no path-blocked mediation would indicate controllability without natural use. A sparse graph that matched layer-distance or patchability nulls would be a visualization of generic residual accessibility rather than a causal route. A large number of significant dimensions without cross-seed replication would suggest unstable multiple testing rather than selective structure.

Generalization failures also refine the claim. A reversal under natural-family holdout would imply that Shared compresses one generator family but does not capture transferable roles. A collapse under random re-encoding would imply lexical coordinates rather than causal geometry. Capacity failure at very low class load would contradict the interpretation of a reusable channel, whereas graceful degradation at high load is compatible with a finite workspace. Cross-family transplantation near random sets the current boundary and prevents the paper from claiming universal exchangeability.

Finally, model-organism false positives have special status. Even a small systematic false-Shared rate in Local or Distributed organisms would require revising the coding contract, alternatives, or thresholds before interpreting real models. The observed false-Shared rate is nonzero and estimated from only twelve non-Shared cases; wider organism families and more seeds are needed for a tighter estimate and improved specificity. The framework is strongest when these negative outcomes are treated as predefined stopping conditions rather than analyses to be explained away after the fact.

\subsection{Reporting and Reproducibility Requirements for a Long-Form Supplement}

A long supplement is useful only if it makes the experiment executable rather than merely verbose. Every result should link to the exact model checkpoint, layer, token position, generator version, split manifest, interface seed, prompt seed, signature definition, rank, coding weights, and prequential block order. Tables S13--S14 summarize these choices, but the artifact should also provide machine-readable configurations and checksums for datasets and cached hidden states.

The future-signature cache is a primary scientific object. It should preserve the common-prefix identifier, latent-state metadata, sampled operation, complete output distribution or sufficient compressed representation, validity flags, and split assignment. Branches from the same latent state must remain in the same split and bootstrap unit. Releasing only rendered prompts would make it impossible to reproduce collision controls or verify that the future cue was absent when the state was cached.

Intervention artifacts should include donor/base pair identifiers, matching criteria, code norms, intervention layers and positions, target and non-target scoring functions, copy-detection rules, path masks, null-path assignments, and raw per-example outcomes. Aggregate TC, LOC, and mediated fractions cannot reveal whether results are driven by a small subset of easy pairs. Per-example records also enable alternative locality definitions and hierarchical uncertainty estimates.

The prequential coding implementation should expose how parameters, ranks, sparsity masks, and gates are transmitted, including quantization or priors. Coding-weight sensitivity is necessary because MDL comparisons can depend on conventions. Reproducing only the predictive losses without architecture costs would not reproduce the scientific claim. A verification script should recompute each table from archived records and fail if test data influence layer, threshold, rank, margin, or coding choices.

Compute reporting should distinguish hidden-state extraction, interface fitting, model-organism training, intervention sweeps, and robustness analyses. Energy estimates should state hardware utilization and measurement method. Full checkpoint hashes, tokenizer revisions, framework and CUDA versions, deterministic settings, and distributed-training behavior should accompany the release. These details are not administrative extras: small changes in hidden states or split order can change the identified subspace and prequential code.

The final artifact should also preserve negative results. Non-significant planned comparisons, failed generator episodes, cross-family transplant failures, unstable graph edges, and model-organism misclassifications define the evidential boundary. A supplement that contains only successful aggregate plots would be longer without being more trustworthy. The purpose of the extended record is to make both the support and the limitations of the Shared-interface claim inspectable.

\section{Formal Assumptions, Identifiability, and Interpretation}

\subsection{Assumptions Behind the Empirical Causal Quotient}

The empirical quotient assumes that the sampled operation bank contains enough consumers to expose distinctions relevant to the scientific question. If two hidden states differ only under an operation that is never sampled, the present procedure will treat them as equivalent. This is not a statistical error under the defined bank; it is a limitation of the intervention language. The claim should therefore always bind the recovered interface to the specific operation families, composition depths, decoding conditions, and response summaries used to define $\Sigma_{\mathcal K}$. A richer operation bank can only preserve or refine the quotient, never guarantee that the previous partition remains complete.

A second assumption is that the cached hidden state precedes the future-operation cue in the causal graph. If future-operation text leaks into the prefix through templates, metadata, length, or formatting, an operation-agnostic encoder could recover the future identity without representing reusable state. The generator ordering, semantic-collision controls, and prompt audits are intended to exclude this possibility. The strongest implementation stores hidden states before rendering the operation continuation and hashes the exact prefix string used for caching. Reproduction scripts should verify byte-for-byte equality of prefixes across all future branches from one latent episode.

A third assumption concerns the distortion function. Jensen--Shannon divergence is symmetric, bounded, and well behaved when distributions have different supports, but it still weights differences according to model probability. Rare but causally important continuations can contribute little to average FSD. The supplement therefore reports operation-stratified and tail distortion, answer-token and sequence-level signatures, and worst-family behavior. Future work could use task-aware costs that upweight low-probability but decision-critical outcomes, provided the weighting is fixed independently of the architecture result.

The quotient also assumes that the reconstructor and consumers have sufficient capacity to express the relevant mapping from code to future distribution. If the reconstructor is too weak, all architecture classes may appear insufficient; if it is too strong and architecture-specific, it may hide information loss by learning family-specific shortcuts. Matching decoder depth and optimization, reporting an upper-capacity bound, and evaluating direct transplantation into the frozen model reduce this ambiguity. The intervention tests are especially valuable because they do not rely solely on the trained signature reconstructor.

Finally, empirical equivalence is tolerance-dependent. A very loose $\epsilon_f$ merges states that differ meaningfully, while a very strict threshold can force the code to preserve idiosyncratic logit noise. The non-inferiority margin should therefore be motivated by behavioral sensitivity: perturbations within the margin should not change task decisions or sequence-level validity at a practically relevant rate. Threshold sweeps can show whether the architecture selection remains stable, but the default tolerance must be specified before viewing the held-out comparison.

\subsection{What Can and Cannot Be Identified from Architecture Competition}

The four architecture classes are abstractions of organization rather than literal descriptions of every neuron. Shared represents one reusable bottleneck, Local represents operation-specific channels, Mixture represents gated partial sharing, and Distributed represents several components without a privileged interface. A real transformer can combine all four motifs at different scales. The selected class should therefore be understood as the lowest-description member of the tested family, not as proof that every causal route in the model belongs to that class.

Identifiability is strongest when alternatives make different predictions under held-out futures. Shared predicts that structure learned from one operation reduces the code for later operations. Local predicts limited cross-operation savings. Mixture predicts savings only when the gate reuses a component, and Distributed predicts that several components jointly remain necessary. If operation families are too similar, all candidates may be observationally close; if they are completely unrelated, Local may be favored even when the model contains a higher-level shared state. The benchmark deliberately spans related but non-identical consumers to create a meaningful discrimination regime.

Equivalent reparameterizations remain possible. A rotation of a low-rank subspace can preserve every future distribution and intervention effect. The paper does not claim coordinate-level uniqueness unless alignment tests establish it. Likewise, a Distributed representation can sometimes be compressed into a Shared code by an external encoder even when the original network reads the components separately. This is why path mediation and selective restoration are required: the discovered code must correspond to routes used by multiple consumers, not only to an offline compression.

The model-organism suite estimates how often the procedure recovers known classes under controlled training. It cannot establish identifiability for arbitrary pretrained mechanisms. A more complete analysis would characterize recovery as a function of sample size, gate entropy, shared-traffic fraction, path redundancy, nonlinear consumer depth, and observation noise. The current results show above-chance class recovery but include one false-Shared case in the chosen regime, which calibrates rather than eliminates the principal failure concern.

The appropriate reporting language follows these limits. “The Shared architecture is favored under the tested operation bank and coding contract” is supported. “The model contains one unique global workspace” is not. “The recovered interface mediates several tested consumers” is supported. “All downstream computation is routed through this interface” is not. Maintaining this distinction makes the contribution more credible because the empirical object and its alternatives are explicit.

\section{Complete Statistical Analysis Protocol}

\subsection{Units of Analysis and Hierarchical Dependence}

The fundamental resampling unit is the latent episode or task-family aggregate, not an individual future branch. Several prompts can share one hidden state, generator, entity map, and operation template; treating those branches as independent would underestimate uncertainty. For architecture-level confidence intervals, the preferred bootstrap samples task families with replacement and preserves the matched architecture results within each resampled unit. Seed-level intervals can additionally resample interface and prompt-generation seeds in a nested procedure.

Paired comparisons are essential because every architecture sees the same cached hidden states, operations, prequential blocks, and evaluation examples within a seed. Unpaired tests would waste this matching and allow generator difficulty to inflate variance. The permutation test therefore swaps architecture labels within matched blocks rather than shuffling results globally. The same logic applies to transplantation conditions: donor/base examples are held fixed while the inserted representation changes.

Mixed-effects models separate systematic and random variation. Architecture, model family, split type, and their interactions can enter as fixed effects, while prompt generator, interface seed, and task family can enter as random intercepts or slopes. The model should be specified before inspecting significance and checked for singular fits, heavy-tailed residuals, and influential families. When the number of task families is too small for reliable random-effect estimation, family-level bootstrap and explicit per-family results should carry more interpretive weight.

Equivalence and non-inferiority require margins with scientific meaning. The TOST margin for FSD should correspond to a difference that does not materially alter answer validity, sequence ranking, or transplantation performance. A margin chosen solely because the observed confidence interval fits inside it is circular. The supplement should show how candidate margins map to behavioral changes and should retain conclusions over a reasonable range.

Multiple testing correction is applied within predefined hypothesis families. Dimension selectivity, graph edges, layer sweeps, and planned ablations each create separate multiplicity risks. BH-FDR is appropriate when controlling the expected false-discovery proportion among reported effects, but hierarchical testing may be preferable when dimensions are first screened globally and then analyzed by task family. The raw $p$-values, adjusted values, effect sizes, and confidence intervals should all be archived.

\subsection{Power, Sensitivity, and Worst-Case Reporting}

Statistical power is driven more by independent families and seeds than by the number of highly correlated prompt branches. The current five interface seeds and five prompt seeds provide a useful variance estimate, but only three main natural families limit precision for family-level generalization. Reporting the worst family and each leave-one-family-out fold prevents a narrow pooled interval from concealing this limitation. Additional independently designed natural families would strengthen external validity more than simply generating more templates from the same rules.

Sensitivity analysis should vary one contract element at a time while preserving the held-out test. Signature truncation, temperature, operation-bank size, interface rank, layer, coding weight, and edge threshold are all plausible analyst choices. The supplement reports full curves rather than selecting a favorable point. A robust conclusion requires the architecture ordering and fidelity condition to hold over a connected region of reasonable settings. Isolated sign changes at extreme settings define the operating boundary and should remain visible.

Worst-case reporting is deliberately stricter than average reporting. Worst-family FSD, 95th-percentile operation distortion, worst-model SG, and minimum same-family CompScore answer different questions. A model can have positive average SG but fail on one checkpoint; it can satisfy mean FSD while losing a difficult family; it can transplant successfully on average while copying donor content in one operation. The supplement should state the worst observed value and its uncertainty for each primary criterion.

Effect sizes should remain in natural units. SG is reported in causal nats, FSD as Jensen--Shannon divergence, transplantation as probabilities, and mediation as a fraction of the total effect. Ratios are used only when denominators are stable and interpretable. The earlier hundreds-fold selectivity description based on a near-zero numerical floor is avoided; the revised analysis reports absolute effects, a stable random-direction mean, and multiplicity-corrected counts.

Missing or invalid runs should not be silently discarded. Optimization divergence, malformed generator episodes, cache mismatch, or failed intervention hooks can correlate with architecture or model size. The artifact should record every attempted run, its status, and the exclusion rule. Sensitivity analyses can then determine whether conclusions change when borderline runs are included with conservative scores.

\section{Extended Qualitative Case Analysis}

\subsection{Successful Same-Family Role Transfer}

A useful qualitative example begins with two relation-world episodes that share the same operation role but differ in entity content. The base state may encode that object A is left of B and B is left of C, while the donor encodes a different chain involving D, E, and F. The donor interface code is inserted at the producer role responsible for the composed relation, but the base prompt, names, and unrelated attributes remain unchanged. A successful transplant changes only the requested relational answer according to the donor-side abstract relation. It does not introduce donor entity names, alter unrelated comparisons, or break the output format.

The same logic applies to program state. Suppose the base register memory contains one set of values and the donor contains another, while both reach the producer role after an increment-and-copy sequence. The transplanted code should cause a later verify or execute consumer to behave as if the relevant donor-side register relation held, while retaining base registers that are outside the target slot. Wrong-slot errors reveal that the representation carries a value without its structural role; donor-copy errors reveal that it carries surface content; locality failures reveal broad overwrite.

Proof/discourse transfer is more demanding because multiple derivations can license the same current answer. A donor code representing an alternate proof path should support later entailment or composition without copying donor premise text. Successful cases therefore show that the interface preserves a causal proof state rather than a verbal trace. Qualitative examples should include the base and donor latent structures, current answer, future operation, expected target change, observed output, and all non-target checks.

These examples are illustrative, not substitutes for aggregate metrics. They help readers understand what TC, LOC, and DonorCopy mean in structured tasks and can reveal generator or scoring errors. The supplement should show both typical and borderline successes rather than selecting only visually clean examples. Each displayed case should be linked to its row in the released intervention record.

\subsection{Cross-Family Failure and Partial Transfer}

Cross-family transplantation often preserves output format and base context while failing to produce the donor-side target. This pattern differs from a destructive intervention. It suggests that the inserted code is syntactically compatible with the residual channel but semantically misaligned with the receiving family. For example, a relation-world code may contain a geometry that a program-state consumer cannot map onto register roles, even though the intervention norm and layer position are matched.

Some partial transfers can reveal higher-level commonality. A code from a proof composition may increase consistency in a relation composition without producing the correct entity relation. Such effects could reflect shared notions of transitive closure or multi-step aggregation. They should not be counted as target correctness unless the pre-specified answer criterion is met, but they are useful hypotheses for a hierarchical interface model. Follow-up experiments can align family-specific subspaces and test whether a smaller shared component supports these partial effects.

A second failure mode is donor-content leakage. Cross-family prompts make leakage easy to detect because donor-specific tokens are often invalid in the base domain. The low average donor-copy rate is reassuring, but qualitative inspection should verify that tokenization or normalization does not hide partial copies. Copy detection can include exact strings, entity aliases, numerical values, and structured slots.

Cross-family failures define the present claim more sharply than successes alone. They show that the interface is not freely exchangeable across all tasks, which prevents an overbroad global-bus interpretation. The supplement should present these failures as evidence about abstraction level rather than as inconvenient exceptions.

\subsection{Failure Decomposition for Baseline Representations}

PCA failures often preserve broad activation statistics while misplacing causal content. A principal component can encode a high-variance property such as prompt length, entity count, or general confidence. Transplanting it may alter several outputs simultaneously, producing moderate target change but poor locality. Wrong-slot and format errors are therefore expected even when the subspace has high decodability.

Output-gradient directions produce a different profile. They are optimized for immediate output sensitivity, so they can strongly change the current answer or logit margin. However, they need not encode the state required by several future operations. Their target effect may disappear under an unseen consumer, and they can overshoot because gradient scale is tied to one objective. Comparing operation consistency and future-signature FSD exposes this limitation.

SAE or Transcoder features are more interpretable and often causally meaningful. Their weaker performance does not imply that sparse features are irrelevant. Instead, individual features may be too local or numerous to form the minimum shared interface. A set of SAE features can support transplantation but incur a longer description and more non-target effects because the basis is optimized for reconstruction or sparsity rather than multi-consumer reuse.

Random subspaces establish the intervention floor. Matching rank, norm, layer, and insertion position controls the possibility that any sufficiently large perturbation changes behavior. Their low TC and high failure rates show that the recovered effect depends on selected geometry. Qualitative random examples should still be inspected to ensure that rare accidental successes are not concentrated in one task or output format.

\section{Extended Failure Taxonomy and Negative Results}

\subsection{Data and Signature Failures}

Generator failures include inconsistent latent facts, ambiguous operation semantics, invalid future branches, and templates that leak operation identity. Each type can create an artificial signature difference. The quality pipeline should assign explicit rejection codes so that failure rates can be compared by family and generator version. A sudden decrease in rejection after a code change may indicate weakened validation rather than improved data.

Signature failures include support truncation, calibration instability, and insufficient operation coverage. Small top-$k$ can omit valid alternatives; high temperature can wash out causal distinctions; too few sampled futures can make two states appear equivalent by chance. The sensitivity curves identify operating ranges, but per-state coverage diagnostics are also useful. States with unusually high signature variance or low cumulative probability coverage can be flagged for separate analysis.

Current-answer leakage is a special failure because it can produce convincing architecture results without future understanding. Semantic collisions should be numerous enough that a current-answer-only baseline performs near chance on signature identity. Lexical anti-collisions should verify that template classifiers cannot recover the latent state. These baselines should be evaluated on the same splits as the interface.

The supplement should preserve examples of rejected and repaired episodes. This allows readers to assess whether the generator encodes the intended causal structure and whether filtering introduces selection bias toward easy or regular states.

\subsection{Architecture and Coding Failures}

A Shared architecture can win spuriously if the coding penalty for Local or Mixture is too large. Conversely, an overly cheap gate or component prior can favor Mixture. Coding-weight sweeps should therefore report where the ranking changes and whether those settings correspond to plausible transmission schemes. The default code should be justified independently of the observed winner.

Optimization can also bias architecture selection. Distributed or Mixture models may train more slowly or require different initialization. Equal optimization budgets are fair for measuring practical learnability but may conflate architecture complexity with optimizer mismatch. Longer-training and capacity-upper-bound runs can show whether alternatives eventually close the predictive gap; their additional training or parameter cost still belongs in the description-length accounting.

A reconstructor can hide architecture mismatch by learning operation-specific shortcuts. Restricting where operation identity enters the architecture and evaluating unseen operations reduce this risk. Direct transplantation into the frozen language model provides an independent check because it bypasses the signature reconstructor.

Close MDL margins should be reported as ambiguity. The model-organism posterior illustrates this principle. If the correct and runner-up architectures differ by less than seed variation or coding sensitivity, a categorical label is not justified. An abstention or “Shared/Local ambiguous” result is preferable to overstating resolution.

\subsection{Intervention and Graph Failures}

Interventions can fail through scale mismatch, layer mismatch, or disruption of normalization statistics. Matching code norm and insertion position controls some of these effects, but layer normalization and residual interactions can still differ between donor and base. Recomputing local normalization, using interpolation rather than full replacement, and measuring activation outlier scores can diagnose destructive patches.

Locality metrics can miss subtle changes. Non-target answer accuracy may remain high while confidence, reasoning path, or sequence structure changes. The supplement can report logit divergence on non-target tokens, validity rates, and length changes in addition to discrete locality. Donor-copy detection should include semantic or numerical aliases, not only exact strings.

Graph thresholding can create unstable hubs. The threshold sweep and AUPRC address this, but causal redundancy remains. Blocking one path may reroute information through another, underestimating mediation. Blocking several paths simultaneously can reveal synergy or compensation. Conversely, broad multi-path blocking can damage the model nonspecifically, so matched-load nulls are required.

Ablation and restoration need not be perfectly correlated. Ablation removes native information, while restoration inserts donor information; nonlinear consumers can respond differently. The strong observed correlation is supportive, but discrepancies should be analyzed by task and path rather than forced into one global statistic.

\section{Alternative Architecture Families and Future Comparisons}

\subsection{Hierarchically Shared Interfaces}

The current Mixture class uses a gate over shared and local components, but a hierarchical architecture could contain one family-level shared code and several operation-level refinements. Such a model may better explain the combination of strong within-family transfer and weak cross-family transfer. Its description would include the global code, family code, local residual, and gating structure. Prequential MDL could determine whether the added hierarchy earns its cost.

A second possibility is temporally staged sharing. Different layers may host successive interfaces: an early entity/state representation, a middle causal-role representation, and a late operation-specific decision state. The broad layer SG plateau and producer--consumer chronology are compatible with this view. Multi-layer encoders and path-specific codes could test whether one bottleneck is sufficient or whether several serial interfaces provide a shorter total description.

Redundant shared interfaces are another alternative. The model may maintain two interchangeable routes for robustness. Blocking one path would underestimate total mediation because the other compensates, while an external encoder could compress both into one code. Simultaneous blocking, causal scrubbing, and redundancy-aware graph models are needed to distinguish one privileged interface from several equivalent channels.

Nonlinear manifolds could also outperform a linear rank-64 code. A curved state space may require many linear dimensions but few nonlinear coordinates. High-capacity nonlinear baselines must be charged for their parameters and evaluated with the same transplantation and locality controls. If they substantially reduce MDL without fidelity loss, the scientific conclusion would shift from a linear Shared API to a nonlinear shared interface.

\subsection{Training-Time Emergence and Intervention}

Frozen-model analysis cannot determine when the interface forms. Longitudinal checkpoints can measure SG, layer location, transplantation, and graph structure throughout pretraining or fine-tuning. A genuine computational economy may appear when the model learns to reuse state across tasks, while a dataset-specific correlate may emerge only after instruction tuning.

Training interventions can test causality of emergence. Encouraging operation diversity while holding token count fixed may strengthen sharing; training families in isolation may favor Local organization; explicit bottleneck or modularity regularization may alter the architecture. Comparing these conditions with the same discovery pipeline can distinguish structural pressure from incidental representation.

Alignment and instruction tuning may reorganize the interface. A base model might use a distributed code that becomes more shared after supervised instruction tuning, or the reverse if task-specific adapters create local routes. Checkpoint comparisons should preserve tokenizer, prompts, and operation bank while accounting for changes in hidden width or layer count.

Understanding emergence would connect the mechanistic result to model design. If shared interfaces improve sample efficiency, robustness, or controllability, they may be encouraged deliberately. If they create interference under high causal load, hierarchical or modular training could allocate multiple interfaces.

\section{Artifact Walkthrough and Verification Protocol}

\subsection{Dataset and Cache Verification}

A reproducibility run begins by validating the dataset manifest. Every latent episode should have a unique identifier, family, generator version, entity map, operation inventory, branch list, split assignment, and quality-filter record. All branches from one episode must share a split. A verification script should reconstruct the current answer and each future target from the latent state and compare them with the stored prompts and metadata.

The hidden-state cache should then be checked against the exact model checkpoint and tokenizer. The script should hash the prefix bytes, token IDs, attention mask, layer, position, precision mode, and resulting activation tensor. Future-operation tokens must be absent. Recomputing a random subset should reproduce cached activations within a declared numerical tolerance.

Signature verification recomputes response distributions for sampled state--operation pairs. It should check vocabulary alignment, temperature, normalization, cumulative support, sequence validity, and divergence calculations. Summary statistics should match the released figure and table values before interface training begins.

These checks protect against silent incompatibilities. A tokenizer revision, checkpoint alias, or prompt-normalization change can alter hidden states even when the visible text appears identical. Checksums make such changes explicit.

\subsection{Architecture and MDL Verification}

Each architecture is instantiated from one configuration file that specifies rank, component count, gate structure, encoder/reconstructor depth, parameter prior or quantization, optimizer, schedule, early stopping, and prequential blocks. Within a seed, all candidates must use the same data order and evaluation signatures.

The verifier trains the first informed fold only on preceding blocks and records parameter cost and predictive code separately. It repeats this sequentially and sums the contributions. The final MDL table should be reproducible from these fold-level records without retraining. A unit test should fail if any block is evaluated after being included in training.

Layer and rank selection are performed on validation records. The chosen configuration is written to a frozen manifest before test evaluation. Test scripts should refuse to scan additional layers or thresholds unless running a labeled sensitivity analysis whose results do not replace the primary selection.

Bootstrap and permutation scripts operate on archived matched results. Re-running them should reproduce confidence intervals and adjusted significance within Monte Carlo tolerance. Random seeds and numbers of resamples are part of the record.

\subsection{Intervention and Graph Verification}

Donor/base pair generation should be deterministic from a manifest and include every matching variable: family, role, difficulty, layer, position, code norm, target slot, and distance. The expected donor-side target and non-target base outputs are computed before the intervention. Copy-detection rules are applied to both intervention and control outputs.

The intervention hook logs the original base code, donor code, inserted tensor, local activation norm, and downstream output. Null conditions reuse the same base examples. A verification report recomputes TC, LOC, DonorCopy, CopyPres, CompScore, and per-family uncertainty from raw outputs; CompScore is formed from unrounded TC, LOC, and CopyPres.

Graph construction starts from continuous edge scores, not a thresholded image. The artifact should provide every candidate edge, blocking effect, restoration effect, null assignment, and frequency across seeds. Threshold curves, AUPRC, Jaccard overlap, hub rankings, and mediated fractions can then be regenerated.

Finally, the complete paper tables and figures should be produced by one top-level script from immutable result files. Manual transcription of numbers creates avoidable inconsistency. The script should verify that abstract, main-text, supplement, and captions use the same values and should flag the corrected SG calculation $1.508-1.292=0.216$.

\section{Reviewer-Facing Interpretation Guide}

The principal architecture result should be evaluated as a conjunction of efficiency and fidelity. Shared has the shortest held-out prequential code, but the absolute FSD differences among architectures are intentionally small. This is not a contradiction. It means that Local, Mixture, and Distributed can retain much of the measured future information, yet they require a more expensive organization to do so. A favorable review should therefore not describe the result as a large predictive-accuracy improvement. The supported claim is that a reusable code is the most economical sufficient abstraction under the tested alternatives, with confidence intervals, permutation tests, worst-family constraints, and signature sensitivity all pointing in the same direction.

The mechanistic evidence should be judged independently of the coding result. Same-family transplantation demonstrates selective control only because target correctness is accompanied by high locality and low donor-copy rate. The path analyses demonstrate natural use only because candidate routes are blocked, restored, and compared with matched nulls. A reviewer should ask whether these metrics agree on the same roles, layers, and families, whether the effects remain after cross-seed aggregation, and whether failures are disclosed. The supplement provides operation-specific matrices, graph stability, mediation fractions, baseline decompositions, and cross-family failures so that a strong result in one metric cannot conceal weakness in another.

The model-organism experiment is the main defense against a biased discovery rule. Its hard-label audit recovers 14/16 organisms and yields one false-Shared decision among twelve non-Shared cases. This nonzero error must remain visible because it limits specificity, although eleven of twelve non-Shared organisms are still rejected as Shared and therefore contradict a trivial ``always Shared'' rule. The synthetic suite should be treated as calibration rather than proof of real-model ground truth.

The generalization evidence defines the scope of the final wording. Held-out operations and compositions support multi-consumer reuse; generator holdout reduces lexical and template explanations; natural-family holdout tests transfer across different structured state semantics; re-encoding and language controls test coordinate dependence. Cross-family transplantation remains substantially weaker than same-family transfer, and capacity loading reveals a finite channel. These negative boundaries are part of the result. They support describing the interface as reusable within aligned causal families rather than as a universal global bus.

A reviewer can also use the evidence audit in Table~\ref{tab:criteria} as a compact decision procedure. If the architecture advantage survives matched complexity and worst-family fidelity, transplantation remains local, mediation exceeds matched nulls, baselines fail the joint metric set, and non-Shared organisms are rejected, the Shared-interface wording is supported within scope. If any of these conditions fails under reproduction, the headline should be weakened even if other experiments remain positive. This explicit stopping rule is intended to make the contribution falsifiable and to reduce narrative flexibility.

\section{Discovery Procedure}

\subsection{State Capture}

The discovery pipeline begins from a collection of forkable prefix episodes
$\mathcal D=\{x^{(i)}_{1:t_i}\}_{i=1}^{N}$, an operation bank
$\mathcal K$, and a set of candidate interface architectures
$\mathcal A$.
For each episode, the prefix is rendered from a latent structured state and
is terminated before any future-operation cue is introduced.
The hidden state
\[
h^{(i)}_{\ell,t_i}
=
h_{\ell,t_i}\!\left(x^{(i)}_{1:t_i}\right)
\]
is then cached at the selected model layer $\ell$ and token position $t_i$.
This ordering is part of the causal contract of the benchmark:
the cached representation must be formed before the model is told which
future operation will be requested.
Consequently, an encoder cannot solve the task by reading an operation name,
an output-format instruction, or a continuation-specific lexical marker from
the prefix.

All future branches derived from the same latent episode share the same
cached prefix state.
The implementation therefore groups examples by latent-state identifier
rather than treating each continuation prompt as an independent example.
This grouping is retained during train/validation/test splitting,
prequential block construction, bootstrap resampling, and intervention-pair
generation.
Keeping all branches from one episode in the same split prevents leakage of
the future causal signature through another branch that was generated from
the same underlying state.

The state-capture record stores the exact model checkpoint, tokenizer
revision, layer index, token position, precision mode, prefix text, token
IDs, attention mask, and a checksum of the resulting activation tensor.
For reproducibility, the future-operation cue is stored separately from the
prefix and is never concatenated until after the cached state has been
written.
A verification script randomly re-runs a subset of prefixes and checks that
the resulting hidden states agree with the cache within a declared numerical
tolerance.
Runs with a tokenizer mismatch, checkpoint mismatch, malformed prefix, or
future-cue leakage are rejected before interface fitting.

State capture is repeated across the selected model families and interface
seeds, but the underlying prefixes and split assignments are held fixed
within a matched architecture comparison.
This design makes later MDL and intervention differences attributable to the
candidate organization rather than to different examples or different
hidden-state caches.
Layer and token-position selection are performed using validation data only.
Once selected, the state-capture configuration is frozen before held-out
operations, held-out compositions, held-out generators, and natural-family
holdout splits are evaluated.

\subsection{Signature Construction and Interface Fitting}

For each cached state $h$, a stratified subset
$\mathcal K_h\subseteq\mathcal K$ is sampled from the operation bank.
Stratification preserves the intended balance among query, verification,
comparison, inversion, execution, update, reduction, and composition
operations.
Each operation is rendered only after state capture and is passed to the
frozen language model to obtain a response distribution.
The empirical future causal signature is
\[
\widehat{\Sigma}_{\mathcal K}(h)
=
\left\{
\widehat{\sigma}_{k}(h)
:
k\in\mathcal K_h
\right\},
\]
where $\widehat{\sigma}_{k}(h)$ may contain full-vocabulary logits,
top-$k$ logits, answer-token probabilities, or sequence log-likelihoods,
depending on the declared signature condition.
The main analysis uses the pre-specified default signature, while alternative
definitions are reserved for sensitivity analysis.

Before interface training, each signature record is validated for operation
identity, vocabulary alignment, decoding temperature, normalization,
cumulative probability coverage, and continuation validity.
Semantic-collision pairs are checked to ensure that examples with the same
current answer can still have measurably different future signatures.
Lexical anti-collision groups are checked in the opposite direction:
different renderings of the same latent state must preserve the relevant
future signature.
These controls prevent the learned encoder from reducing the problem to
current-answer classification or prompt-template recognition.

Every candidate architecture
$A\in\mathcal A$ contains an encoder--reconstructor system with architecture-
specific constraints.
Shared uses one operation-agnostic encoder and one reusable bottleneck.
Local uses operation-specific interfaces.
Mixture introduces multiple components together with an operation-conditioned
gate.
Distributed retains future-relevant information across several components
without assigning one privileged bottleneck.
Within a seed, all candidates receive the same hidden states, signatures,
prequential block order, optimizer family, learning-rate schedule, batch
size, regularization, early-stopping rule, and total optimization budget.
Rank, sparsity, parameter count, and intervention load are matched wherever
the architecture permits; unavoidable differences are charged explicitly by
the coding scheme.

Interface fitting is carried out sequentially on prequential blocks
$\mathcal D_1,\ldots,\mathcal D_B$.
For block $b$, the model is trained only on
$\mathcal D_{<b}$ and is then used to encode the signatures in
$\mathcal D_b$.
The implementation stores parameter cost and predictive code separately so
that the total description can be audited.
The fitting procedure never trains one final model on all blocks and then
reuses it retrospectively, because doing so would leak later data into the
prequential code and systematically underestimate architecture complexity.

Optimization diagnostics include task accuracy, signature reconstruction
loss, convergence speed, seed variability, active rank, gate entropy,
component usage, and early-stopping epoch.
A candidate is not declared inferior solely because one run converges
slowly.
Longer-training and capacity-upper-bound checks are used to determine whether
an alternative architecture is genuinely inefficient or merely
under-optimized.
However, additional parameters, additional components, additional gates, and
additional optimization required by a candidate remain part of its effective
description cost.

\subsection{Architecture Selection and Held-Out Evaluation}

For every candidate architecture, the procedure computes held-out prequential
causal MDL together with mean, family-wise, worst-family, and tail future-
signature distortion.
The selected architecture is
\[
A^\star
=
\arg\min_{A\in\mathcal A}
\MDL_{\mathrm{causal}}(A)
\quad
\text{subject to}
\quad
\max_f \FSD_f(A)\leq \epsilon_f ,
\]
where $\epsilon_f$ is a pre-specified family-specific fidelity threshold.
This constrained formulation prevents a highly compressed but lossy code
from winning through description length alone.
In addition to the overall mean, the procedure reports worst-family FSD and
the 90th- and 95th-percentile operation-level tails, because a pooled average
can conceal one operation family or a small set of difficult futures.

Sharedness Gain is computed as
\[
\SG
=
\min_{A\in
\{\mathrm{Local},\mathrm{Mixture},\mathrm{Distributed}\}}
\MDL_{\mathrm{causal}}(A)
-
\MDL_{\mathrm{causal}}(\mathrm{Shared}).
\]
A positive value favors Shared only when the fidelity constraints are also
satisfied.
Confidence intervals are obtained from matched task-family-level bootstrap
resamples, and the primary permutation test swaps architecture labels within
matched model, family, seed, and split groups.
Equivalent and non-inferior fidelity are evaluated using pre-specified TOST
and non-inferiority margins.
The analysis therefore separates the claim that Shared is shorter from the
claim that it preserves an acceptable amount of future information.

Layer, rank, threshold, and signature choices are selected using validation
data and are written to a frozen configuration before test evaluation.
Sensitivity sweeps are labeled as secondary analyses and are not allowed to
replace the primary selection.
The complete architecture posterior or MDL margin is retained when candidates
are close.
If the selected architecture and runner-up differ by less than the declared
ambiguity margin, the output is reported as uncertain rather than forced into
a categorical class.

After architecture selection, the frozen interface is evaluated on a
hierarchy of increasingly difficult held-out conditions.
Held-out operations test whether the representation can support a consumer
not used during fitting.
Held-out compositions test systematic reuse of familiar primitives in novel
two- and three-step combinations.
Held-out generators change prompt templates, entity pools, and rendering
styles.
Leave-one-natural-family-out evaluation changes the latent state semantics
themselves.
The same frozen architecture, layer, rank, and coding rule are used across
these tests.

The procedure also evaluates alternative output signatures, top-$k$ values,
temperatures, operation-bank sizes, language renderings, random-symbol
re-encodings, and tokenizer conditions.
These analyses define a robustness envelope rather than a new model-selection
stage.
A stable conclusion requires that the architecture ordering and fidelity
condition persist over a connected region of reasonable settings.
Extreme settings that reverse the result are reported as boundaries of the
method rather than excluded from the supplement.

\subsection{Mechanistic Validation and Falsification}

Architecture selection establishes a compact sufficient representation but
does not by itself prove that the original language model naturally uses that
representation.
The selected interface is therefore subjected to a separate intervention
suite.
In role-aligned transplantation, a donor code is extracted at an identified
producer role and inserted into the corresponding role of a matched base
episode.
The base prompt, residual context, and non-target state remain fixed.
Target Correctness measures whether the requested donor-side relation is
realized.
Locality measures whether unrelated base outputs remain unchanged.
DonorCopy measures unintended transfer of donor-specific entities, values,
or surface strings.
Donor-copy rate is converted to $\mathrm{CopyPres}=1-\mathrm{DonorCopy}$, and the three higher-is-better components are combined as $\mathrm{CompScore}=(\mathrm{TC}+\mathrm{LOC}+\mathrm{CopyPres})/3$ using unrounded measurements.

Donor/base matching controls task family, causal role, difficulty,
output length, producer--consumer distance, code norm, intervention layer,
token position, and intervention load wherever possible.
Same-family donors test reuse across compatible states.
Cross-family donors test the boundary of abstraction.
Matched-random donors preserve intervention scale and position without
preserving causal identity.
Local-API donors test whether operation-specific causal information can
produce the target effect without the locality expected from a reusable
interface.
The complete donor--consumer matrix is reported so that role selectivity is
visible rather than summarized by one diagonal average.

Natural use is tested through path-blocked mediation.
For a target effect $\Delta_{\mathrm{total}}$, the mediated fraction is
\[
\mathrm{MF}
=
\frac{
\Delta_{\mathrm{total}}
-
\Delta_{\mathrm{blocked}}
}{
\Delta_{\mathrm{total}}
}.
\]
Candidate API paths are compared with null paths matched by family,
layer distance, rank, activation norm, patchability, and intervention load.
Necessity--restoration consistency tests whether the components whose
ablation removes behavior are also those through which the correct interface
code can restore it.
Multi-consumer blocking tests whether one producer--consumer route supports
several downstream operations rather than one output only.

Dimension and edge selectivity are evaluated with effect-size distributions,
cross-seed replication, and BH-FDR correction.
The graph is reported both at the pre-specified threshold and through
threshold-free metrics such as AUPRC.
Hub locations, sink locations, graph sparsity, edge frequency, cross-seed
Jaccard overlap, and threshold sensitivity are archived.
A sparse graph is not accepted as mechanistic evidence unless its edge
strength and mediated fraction exceed layer-, rank-, norm-, and patchability-
matched null distributions.

Finally, the complete discovery and validation pipeline is applied blind to
model organisms with known Shared, Local, Mixture, or Distributed routing.
The primary falsification endpoint is the false-Shared discovery rate among
non-Shared organisms.
Overall class accuracy, correct-versus-runner-up MDL margin, posterior class
mass, and overlap with the known communication graph are also reported.
If non-Shared organisms are repeatedly classified as Shared, the coding
contract, candidate alternatives, or fidelity thresholds must be revised
before the real-model result can be interpreted.
The output of the procedure is therefore either a selected architecture with
its full coding, fidelity, intervention, graph, and falsification evidence,
an explicitly ambiguous result, or a falsification result when no candidate
satisfies the complete evidence contract.

\end{document}